\documentclass[format=acmsmall,screen=true,nonacm]{acmart}

\usepackage{etextools}
\usepackage{float}
\usepackage{amsmath}
\usepackage{bbm}
\usepackage{epsfig}
\usepackage{epsf}
\usepackage{psfrag}
\usepackage{verbatim}
\usepackage{color}
\usepackage{multirow}
\usepackage{tabularx}
\usepackage{booktabs}
\usepackage[tight,footnotesize]{subfigure}
\usepackage{array}
\usepackage{soul}
\usepackage{footnote}
\usepackage{dblfloatfix}
\usepackage{tcolorbox}

\usepackage{color, colortbl}
\usepackage{graphicx}
\graphicspath{{Figures}}
\DeclareGraphicsExtensions{.pdf,.png}
\usepackage{bigstrut}
\usepackage[english]{babel}

\usepackage{csquotes}

\usepackage[T1]{fontenc}
\usepackage{algorithm, setspace}
\usepackage{algpseudocode}
\usepackage{url}
\usepackage{multirow}
\usepackage{float}
\usepackage{xcolor}
\usepackage{hyperref}
 \hypersetup{
     colorlinks=true,
     linkcolor=orange,
     filecolor=orange,
     citecolor=orange,      
     urlcolor=orange,
     }
     
\newcommand{\reb}[1]{{\color{black} #1}}

\newcommand{\p}[1]{\smallskip \noindent \textbf{{#1}.}}
\newcommand{\eq}[1]{Equation~(\ref{eq:#1})}
\newcommand{\fig}[1]{Figure~\ref{fig:#1}}
\usepackage{balance}

\usepackage{enumitem}

\def\BibTeX{{\rm B\kern-.05em{\sc i\kern-.025em b}\kern-.08em
    T\kern-.1667em\lower.7ex\hbox{E}\kern-.125emX}}

\setcopyright{none}

\begin{document}

\title{PECAN: Personalizing Robot Behaviors through a Learned Canonical Space}

\author{Heramb Nemlekar}
\orcid{0000-0002-6806-9704}
\email{hnemlekar@vt.edu}

\author{Robert Ramirez Sanchez}
\orcid{0009-0009-0425-5338}
\email{robertjrs@vt.edu}

\author{Dylan P. Losey}
\orcid{0000-0002-8787-5293}
\email{losey@vt.edu}

\affiliation{
  \institution{Virginia Tech}
  \department{Department of Mechanical Engineering}
  \streetaddress{635 Prices Fork Rd}
  \city{Blacksburg}
  \state{VA}
  \postcode{24060}
  \country{USA}
  }

\thanks{This work is supported in part by NSF Grant \#2246446.}

\begin{abstract}

Robots should personalize how they perform tasks to match the needs of individual human users.
Today's robots achieve this personalization by asking for the human's feedback in the task space.
For example, an autonomous car might show the human two different ways to decelerate at stoplights, and ask the human which of these motions they prefer.
This current approach to personalization is \textit{indirect}: based on the behaviors the human selects (e.g., decelerating slowly), the robot tries to infer their underlying preference (e.g., defensive driving).
By contrast, our paper develops a learning and interface-based approach that enables humans to \textit{directly} indicate their desired style.
We do this by learning an abstract, low-dimensional, and continuous canonical space from human demonstration data.
Each point in the canonical space corresponds to a different style (e.g., defensive or aggressive driving), and users can directly personalize the robot's behavior by simply clicking on a point.
Given the human's selection, the robot then decodes this canonical style across each task in the dataset --- e.g., if the human selects a defensive style, the autonomous car personalizes its behavior to drive defensively when decelerating, passing other cars, or merging onto highways.
We refer to our resulting approach as PECAN: \textbf{Pe}rsonalizing Robot Behaviors through a Learned \textbf{Can}onical Space.
Our simulations and user studies suggest that humans prefer using PECAN to directly personalize robot behavior (particularly when those users become familiar with PECAN), and that users find the learned canonical space to be intuitive and consistent.
See videos here: \url{https://youtu.be/wRJpyr23PKI}

\end{abstract}

%
%


%
%


\maketitle

\section{Introduction}
\label{sec:intro}

Over its lifetime, a robot will likely interact with multiple different humans.
Each of these humans has their own personal preferences for how the robot should behave.
For example, consider an autonomous car that drives a human passenger.
One passenger might prefer for the autonomous car to be especially defensive, while another passenger may want the car to drive more aggressively.
In order to account for this personalization, we recognize that it is not sufficient for a robot to just learn the desired tasks it should perform.
We also need robots that adapt the \textit{way they perform those tasks} (i.e., adapt their \textit{style}) to match the current user's preferences.

Existing research often tries to address this problem by collecting human feedback in the task space.
Here the human can demonstrate their desired trajectory, correct the robot's motion, or rank the robot's behavior \cite{nikolaidis2015efficient, ravichandar2020recent, sadigh2017active, zhan2021human, spencer2022expert, losey2022physical, munzer2017preference, biyik2022learning, mehta2023unified, hejna2023few}.
The robot then uses this feedback to update its estimate of what style the human truly wants: for example, if a human passenger shows the autonomous car that it should gradually decelerate at red lights, the autonomous car might infer that the human prefers defensive driving.
Unfortunately, this task-space approach is fundamentally limited because it results in \textit{indirect personalization}.
The human user is not able to directly convey their desired style (e.g., defensive driving). 
Instead, the human must show behaviors that exhibit their desired style in the task space (e.g., gradually decelerating), and hope that the robot infers the correct style from this data.

\begin{figure}[t]
    \begin{center}
    \includegraphics[width=0.6\columnwidth]{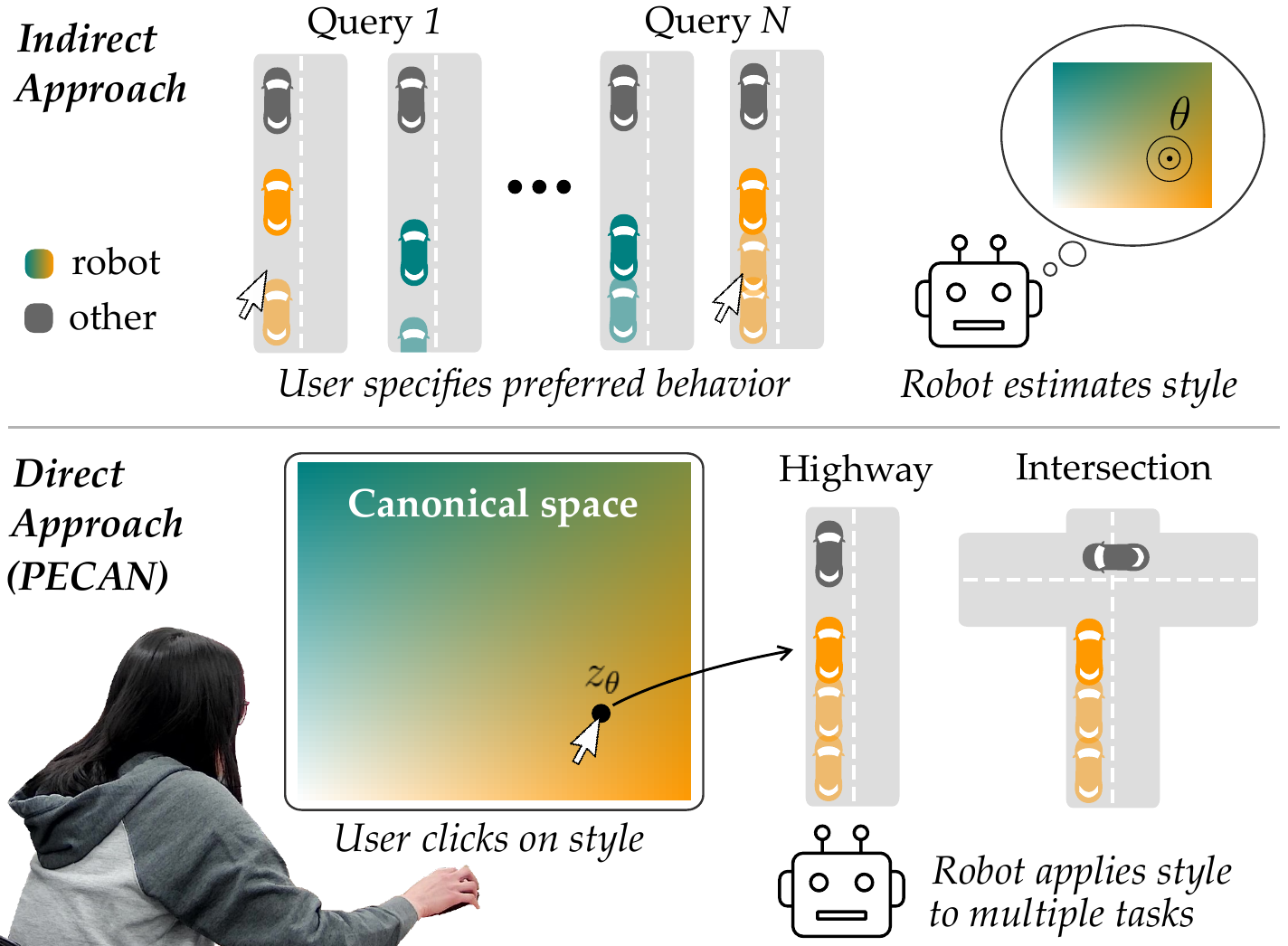}
    \caption{User personalizing the driving style of an autonomous car. With existing methods, the user provides feedback about their preferred behavior, and the robot \textit{indirectly estimates} their style based on this feedback. We propose a \textit{direct} approach where users select their style in a canonical space, and the robot applies this user-chosen style across each task it encounters.}
    \label{fig:front}
    \end{center}
\end{figure}

By contrast, in this paper we enable humans to \textit{directly personalize} the robot's behavior across multiple tasks with a single click.
We achieve this shift by lifting human-robot interaction from the task space into the style space. 
More specifically, we hypothesize that:
\begin{center}
\textit{Styles are often shared across tasks, and these \\ common styles can be captured by a \emph{canonical space}}.
\end{center}
We leverage our insight to introduce \textbf{PECAN}: \textbf{Pe}rsonalizing Robot Behaviors through a Learned \textbf{Can}onical Space.
At training time, this algorithm collects human demonstrations of diverse tasks (e.g., decelerating at a red light, yielding at an intersection).
The robot then leverages our proposed representation learning approach to autonomously extract a high-level canonical space from the task demonstrations.
This learned canonical space is an abstract but user-friendly manifold: each point in the manifold corresponds to a different underlying style.
At run time, humans select a point from this canonical space (i.e., each user can choose their own desired style), and the robot decodes that point into consistent behaviors across each task in the dataset.
For example, if the current passenger selects a point that corresponds to defensive driving, the autonomous car will apply this canonical style to each individual task: decelerating gradually at stop lights, staying far from other vehicles on the highway, and yielding the right of way at intersections.

Overall, PECAN enables humans to rapidly personalize the way a robot behaves by directly specifying their preferred style. We make the following contributions:

\p{Learning a Canonical Space of Styles}
We introduce an approach for learning a high-level representation of robot behaviors through weak supervision. Given data of diverse human demonstrations and a few labels for demonstrations with similar styles, the robot encodes information about the tasks and styles into separate spaces. This allows users to choose their preferred style independent of the tasks. 

\p{Forming a User-Friendly Canonical Space}
We propose characteristics of the canonical space --- such as consistency and monotonicity --- that make it easy for users to understand how the styles are encoded and select their preferred style. The robot induces these characteristics by using demonstrations that represent the extremes of the style spectrum.

\p{Evaluating with Real Users}
We compare our approach to state-of-the-art baselines in two user studies: one using a robot arm and another focusing on an autonomous driving scenario. Our results suggest that participants find it easier to personalize robot behaviors using an interface that leverages our approach. They also prefer PECAN over the baselines, stating that it is more consistent and intuitive. These results become more pronounced as the users gain experience using PECAN, suggesting that familiarization is an important factor in our method's effectiveness.
\section{Related Work}

\noindent \textbf{Preference Learning.}
Humans can personalize a robot by providing feedback about its behaviors. 
This includes demonstrating the desired motion~\cite{nikolaidis2015efficient, ravichandar2020recent}, correcting the robot's actions~\cite{li2021learning, spencer2022expert, losey2022physical}, selecting their preferred trajectory from options presented by the robot~\cite{sadigh2017active, wilde2020improving, zhan2021human}, or some combination of these feedback types~\cite{jain2015learning, munzer2017preference, biyik2022learning, mehta2023unified}. 
These works parameterize the robot's behavior with user-defined features, such as the speed of an autonomous car or its distance from other cars in a driving scenario. The parameters or weights for these features are then estimated based on the human's feedback. \reb{Together, the weights and features produce the user's preferred behavior. We refer to this resulting behavior as the robot's \textit{style} (i.e., how it performs the given task)~\cite{rosbach2019driving}}. 

Specifying the correct set of features for modeling complex styles can be challenging~\cite{bobu2024aligning}. 
\reb{
While recent approaches have explored learning the features online~\cite{katz2021preference}, humans still need to teach new features by providing further demonstrations~\cite{bobu2022inducing}.}
To overcome this, previous research has also proposed deep reinforcement learning from human feedback~\cite{christiano2017deep, hejna2023few}, which bypasses the need to specify such features. 
However, this approach is often time-consuming, as users may have to provide feedback hundreds or even thousands of times throughout the learning process.

Our work is different from these approaches in two ways. First, we do not pre-define the styles or features. Instead, we learn a latent representation of the styles from offline human demonstrations. Second, we do not collect more data from users in the task space. Instead, we design a canonical space that allows users to select their preferred style by simply clicking on the learned representation.


\p{Multi-task Personalization}
The methods discussed so far involve a single robot model that learns both the task and the user's preferred way of performing it.
However, our insight is that these style preferences are often shared across diverse tasks. For instance, humans tend to prefer similar navigation styles across different search and rescue scenarios~\cite{guo2021transfer}. Some recent methods account for this by separately learning the user styles from the task-specific objective~\cite{amin2017repeated, woodworth2018preference, bonyani2024style}. However, these works still assume that the relevant features are known. Our approach will also separate the learning of user styles from the tasks, but without assuming any features.

Specifically, we propose to construct a \textit{canonical} representation of styles that is shared across multiple tasks. 
Previous research in multi-task learning has similarly explored how robots can learn canonical representations over several tasks~\cite{rahmatizadeh2018vision, singh2020scalable, allshire2021laser, mandi2022towards}. For example, in 
\cite{rahmatizadeh2018vision, mandi2022towards}, the robot learns a common visual representation of various tasks, and in \cite{allshire2021laser}, the robot learns a latent action representation that applies to a family of tasks. Our approach differs from these works because --- in addition to learning multiple tasks --- we also learn a canonical space that captures the different styles with which these tasks can be performed.


\p{Learning a Canonical Style Space}
Most relevant to our approach are methods like \reb{\cite{lynch2020learning, osa2020goal,jiao2022tae, xihan2022skill}} that learn a canonical representation of robot skills or styles over several tasks. 

In~\cite{lynch2020learning}, the robot embeds action sequences into a latent space of task-agnostic skills. The robot then executes these skills on a continuous range of tasks that are specified by their start and end states. When the set of tasks is finite, \cite{osa2020goal} learns a latent space comprised of discrete and continuous portions. The discrete variables capture the tasks and the continuous variables capture the latent styles. \reb{Similar to other works in embedding robot trajectories~\cite{co2018self, allshire2021laser, sun2021task, wang2022skill}, b}oth these methods employ Variational Autoencoders (VAEs) to train the latent spaces in an unsupervised manner. However, \cite{lynch2020learning} assumes that the tasks are specified by the user, while \cite{osa2020goal} does not guarantee that 
the respective latent spaces will exclusively capture the tasks and styles present in the data.

To address this, \reb{\cite{jiao2022tae} uses labels that specify the intention (i.e., task) and aggressiveness (i.e., style) of an autonomous car's trajectories. Recent work~\cite{schrum2024maveric} has also explored adjusting the aggressiveness of autonomous driving by obtaining subjective ratings of the driving style to learn a tunable latent dimension. However, it is often challenging for humans to precisely quantify the robot's style~\cite{koppol2021interaction}. For this reason, \cite{xihan2022skill} instead requires users to label trajectories that belong to the same task and trajectories that correspond to the same skill}. These labels are used to train Gated VAEs~\cite{vowels2020gated} which encode the task and skill knowledge into separate latent spaces. 
A major limitation of all these methods is that the encoded styles are not consistent across tasks. 
The same latent value can produce different styles depending on the task --- making the interface unintuitive for humans.
Ideally, users should be able to select their preferred style with a single click and produce similar robot behaviors across all tasks. 

Our problem also bears similarities with the style transfer problem in computer vision~\cite{dupont2018learning, karras2019stylegen, jing2019neural, abdal2019image2stylegan, smieja2020segma, rangwani2023noisytwins}. For instance, \cite{karras2019stylegen} and \cite{abdal2019image2stylegan} transfer styles such as hair color and facial expressions from one image to another. However, this means that they require a reference style to copy from, which in our case would be a reference robot trajectory that users would have to demonstrate. In contrast, \cite{smieja2020segma} and \cite{rangwani2023noisytwins} generate styled images by taking a latent class and a latent style value as input. Both approaches encode classes (i.e., tasks) and styles within a single continuous latent space. Specifically, \cite{smieja2020segma} learns a Gaussian mixture representation where each Gaussian represents a class and its variance captures the styles. 
Most similar to ours is Joint VAE~\cite{dupont2018learning}, which learns separate latent spaces for tasks and styles, but in an unsupervised manner.
In this work, we evaluate whether these approaches apply to robotics tasks, comparing our method to \cite{smieja2020segma} and \cite{dupont2018learning}.

Overall, similar to approaches like~\cite{osa2020goal}, we utilize separate discrete and continuous latent spaces to encode the task and style information. The latent style space becomes our canonical space. To ensure that the canonical space is consistent across tasks, we use a small set of labels for trajectories having similar styles but in different tasks. Unlike \cite{xihan2022skill} and \cite{smieja2020segma}, our approach does not require any task labels. Instead, we capture the actual tasks and styles with their respective latent spaces by only using labels for trajectories with similar styles. 

\section{Problem Statement}
\label{sec:problem}

We explore how robotic systems (such as robot arms or autonomous vehicles) can learn a canonical space for personalizing their behaviors.
We assume that the robot is given a dataset with demonstrations of diverse tasks and ways of performing those tasks.
From this dataset the robot needs to extract a low-dimensional and user-friendly manifold that embeds the \textit{styles} shared across tasks (e.g., driving an autonomous car defensively or aggressively).
Importantly, we do not assume that the styles are predefined or that the tasks are known.
Instead, the system must learn these underlying styles to autonomously construct the canonical space.

\p{Trajectories}
Let $s \in \mathcal{S}$ be the system state and let $a \in \mathcal{A}$ be a robot action.
For example, in our driving scenario the state $s$ includes the position and heading of the autonomous car and any other nearby vehicles, and $a$ is the robot's steering and acceleration inputs.
A trajectory $\xi \in \Xi$ is a sequence of $T$ state-action pairs: $\xi = \{(s_1, a_1), \ldots (s_T, a_T)\}$.
We obtain trajectories by rolling out the robot's learned behaviors in the environment, or by collecting demonstrations from humans.

\p{Dataset}
At training time the robot is given a dataset with $N$ demonstrations from one or more human teachers.
Each demonstration is a trajectory, so that the dataset consists of: $\mathcal{D} = \{\xi_1, \ldots, \xi_N\}$.
The trajectories within $\mathcal{D}$ show examples of multiple \textit{tasks}, and perform those tasks with a variety of different \textit{styles}.
Let $\tau \in \mathcal{T}$ be the space of tasks and let $\theta \in \mathbb{R}^{d_{\theta}}$ be the space of styles.
We assume that there are a discrete set of tasks (e.g., slowing for a red light, passing on the highway, crossing an intersection), but the manifold of styles is continuous.
Returning to our driving example, the human can provide trajectories that slow for a red light (i.e., the task) along a spectrum from very gradually to very abruptly (i.e., the style).
Overall, each demonstration $\xi \in \mathcal{D}$ corresponds to some task $\tau$ and style $\theta$.

\p{Labels}
In practice, however, we \textit{do not assume} that the robot knows the task or style for any trajectory $\xi \in \mathcal{D}$.
This is partially because it is difficult for humans to quantify the style of their demonstrations~\cite{koppol2021interaction}.
Imagine a person showing an autonomous car how to smoothly slow down for a red light; what numerical value of $\theta$ should the human give to that behavior?
Rather than asking humans to provide $\theta$, we instead ask users to label trajectories that have \textit{similar styles}.
For our driving example, perhaps in $\xi_1$ the autonomous car brakes late for a red light, and in $\xi_2$ the autonomous car tailgates directly behind another vehicle. 
A human teacher might label $\xi_1$ and $\xi_2$ as having similar styles, since in both trajectories the robot drives aggressively.
More generally, it is up to the human teacher(s) to decide what the styles are, and what groups of trajectories they think have similar styles.
As a result of this process the robot is given labels $\mathcal{Y}$.
Each label $y_{i} \in \mathcal{Y}$ contains a set of trajectories $y_{i} = \{\xi_{1}, \xi_{2}, \ldots\}$ that all have similar styles (as determined by the human teachers).
Every trajectory $\xi \in y$ belongs to the dataset $\mathcal{D}$; however, not all trajectories $\xi \in \mathcal{D}$ need to be labeled in $\mathcal{Y}$.

Overall, the robot's objective is to leverage the dataset $\mathcal{D}$ and labels $\mathcal{Y}$ to learn a canonical space of styles that allows users to easily personalize the robot's behavior across tasks.
\begin{figure*}[t]
    \begin{center}
        \includegraphics[width=1\textwidth]{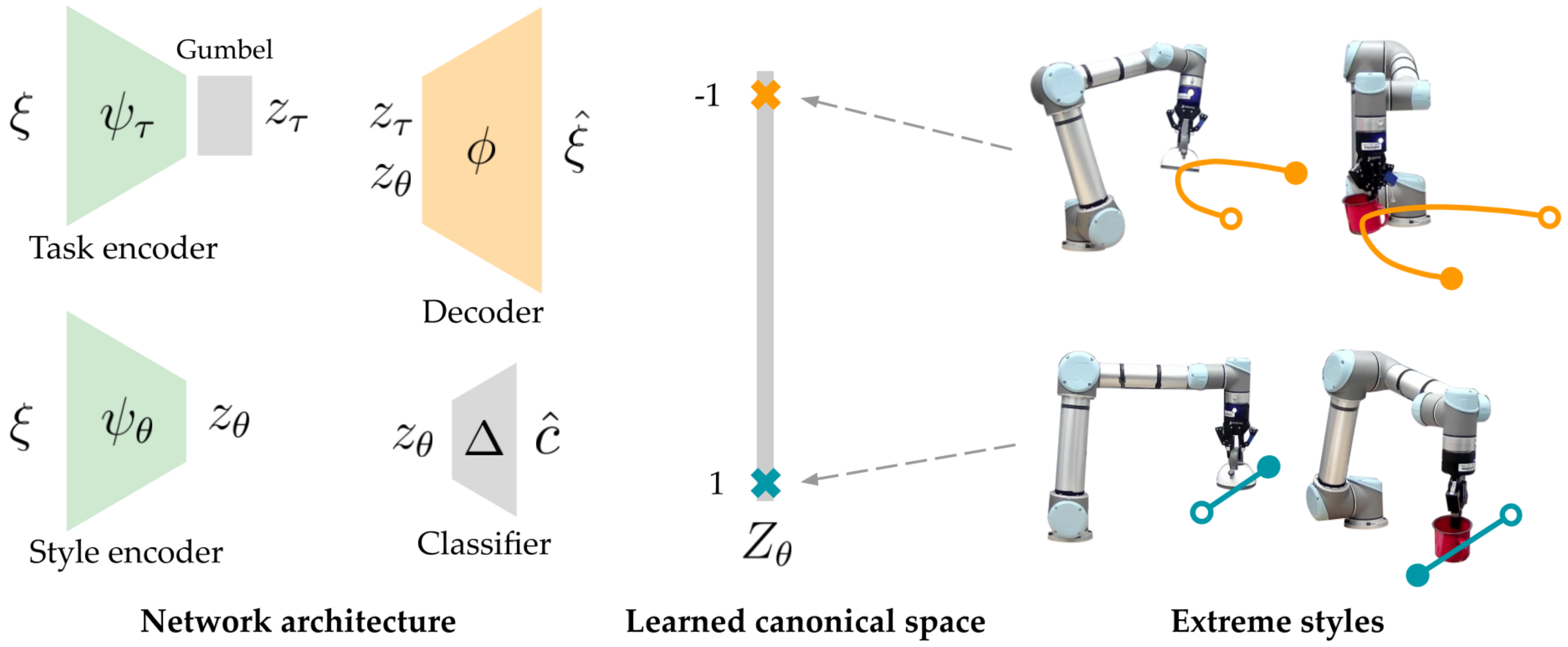}
        \vspace{-3ex}
        \caption{Proposed architecture for \textbf{Pe}rsonalizing Robot Behaviors through a Learned \textbf{Can}onical Space (PECAN). (Left) The robot uses a task encoder $\psi_{\tau}$ and a style encoder $\psi_{\theta}$ to embed input demonstrations $\xi \in \mathcal{D}$ into two low-dimensional spaces: a latent task space $Z_{\tau}$ and a latent style space $Z_{\theta}$. A decoder network $\phi$ takes the combined latent tasks and styles as input and reproduces the input demonstrations. For labeled demonstrations, a classifier network $\Delta$ predicts the class labels from their latent styles. We train both the encoders and the decoder to accurately reconstruct the demonstrations. Simultaneously, we also train the style encoder along with the classifier such that it assigns similar latent values to trajectories with the same label. (Right) We show that when labeled demonstrations represent the extreme ends of the style spectrum, the canonical space is organized so that the latent styles of these extremes are positioned at the corners. This arrangement allows users to interpolate between the extremes by choosing intermediate latent values.}
        \label{fig:method}
        \vspace{-2 ex}
    \end{center}
\end{figure*}

\section{Learning a Canonical Style Space}
\label{sec:method}

We want to enable people to select their preferred style $\theta$ for completing a collection of tasks $\mathcal{T}$ with just a few clicks. 
This personalization is challenging because the robot has no explicit knowledge of either the tasks or styles. 
However --- recalling our motivating hypothesis --- we recognize that styles are often shared across tasks, and so we can try to capture these underlying styles with a learned \textit{canonical space}.

In this section we outline our approach for constructing this canonical space (see \fig{method}). First, in Section~\ref{sec:method-encode}, we introduce an autoencoder architecture that extracts the task and style information from the demonstrated trajectories. Our architecture encodes the tasks and styles into distinct latent spaces; the latent style space becomes our canonical manifold. Next, in Section~\ref{sec:method-characteristics}, we propose characteristics that make the canonical space user-friendly, so that humans can easily interact with that space to search for and select their preferred styles. Finally, in Section~\ref{sec:method-learn}, we describe our training process, focusing on how we learn a canonical space that effectively captures the styles in dataset $\mathcal{D}$ while also exhibiting user-friendly characteristics.

\subsection{Separately Encoding Tasks and Styles} \label{sec:method-encode}

As defined in Section~\ref{sec:problem}, each robot trajectory corresponds to a specific task $\tau \in \mathcal{T}$ and style $\theta \in \mathbb{R}^{d_{\theta}}$. For example, a robot arm could perform tasks like placing a cup in front of the user or pouring coffee into that cup. Different users may prefer different styles for these tasks (see \fig{method}): perhaps one user provides demonstrations where the robot follows the shortest path, while another user shows demonstrations that take an exaggerated path to maintain a safe distance from the human.

Our insight is that these underlying styles are often consistent across tasks. We therefore want to learn a style space that is \textit{independent} of the tasks, so that users can select their preferred style from this space and obtain the corresponding robot trajectory across each task. To facilitate this, we propose an autoencoder architecture with two encoders: a \textit{task encoder} $\psi_{\tau}$ and a \textit{style encoder} $\psi_{\theta}$, as well as one \textit{trajectory decoder} $\phi$. 

\p{Task Encoder} 
The task encoder maps input trajectories $\xi \in \Xi$ to a latent space of tasks $Z_{\tau}$.
$$\psi_{\tau}: \Xi \mapsto Z_{\tau}$$
This latent space encodes the tasks present in our dataset. We assume that data consists of a finite number of discrete tasks $\mathcal{T}$. To capture these tasks, we want the latent task space to also be discrete such that each $z_{\tau} \in Z_{\tau}$ corresponds to a task $\tau \in \mathcal{T}$. Therefore, we discretize the output of the task encoder by applying the Straight-Through Gumbel Estimator~\cite{jang2017categorical}. This technique constrains $z_{\tau}$ to be a one-hot vector of dimensions $d_{\tau} = |\mathcal{T}|$. For instance, in the example shown in \fig{method}, the task of placing the cup could be mapped to $z_{\tau} = [0, 1]$ while the task of pouring coffee is mapped to $z_{\tau} = [1, 0]$.
In our experiments we will assume that the number of tasks $|\mathcal{T}|$ is known, but unsupervised metrics such as Normalised Mutual Information (NMI) can also be used to autonomously estimate the number of discrete tasks in the dataset~\cite{zolotas2022disentangled}.

\p{Style Encoder} 
The style encoder $\psi_{\theta}$ maps input trajectories $\xi \in \Xi$ to a latent space of styles $Z_{\theta}$. We will refer to this latent style space as our \textit{canonical space}:
$$\psi_{\theta}: \Xi \mapsto Z_{\theta}$$
Unlike the discrete space of tasks, we recognize that the robot styles can be continuous. For example, the robot trajectory in \fig{method} can vary along a continuous spectrum from straight goal-directed paths to exaggerated motions that stay far from the human. Accordingly, our canonical space $Z_{\theta}$ is a continuous manifold. 
In our experiments we use a $Tanh(\cdot)$ activation function at the final layer to bound the canonical space within $[-1, 1]^{d_{\theta}}$, so that our resulting canonical space is a $d_{\theta}$-dimensional cube.

\p{Trajectory Decoder} 
Our goal is to let users choose a latent style $z_{\theta}$ from the canonical space and have the robot perform trajectories aligned with that style in each task $z_{\tau} \in Z_{\tau}$.
To achieve this, we include a decoder network $\phi$ that takes both the task and style encodings as input and reconstructs the corresponding robot trajectories: 
$$\phi: Z_{\tau} \times Z_{\theta} \mapsto \Xi$$ 
We train the decoder to accurately reconstruct a trajectory $\xi$ given the values for its latent task $z_{\tau}$ and latent style $z_{\theta}$ by minimizing the following loss:
\begin{equation}
    \mathcal{L}_{trajectory} = \sum_{\xi \in \mathcal{D}} || \, \xi \, - \, \phi (\psi_{\tau}(\xi), \, \psi_{\theta}(\xi) ) \, ||^{2}
\end{equation}
When this loss is minimized, it indirectly encourages the task and style encoders to capture sufficient representations of the trajectories in the dataset. 

\reb{
\p{Implementation}
In our experiments, we model all networks as multilayer perceptions (MLPs) with four fully connected layers and rectified linear unit (ReLU) and hyperbolic tangent (Tanh) activation functions. 
We flatten each trajectory into a vector with the same size as the first layer of the encoder MLPs and the last layer of the decoder MLP. For tasks with high-dimensional states, we first downsample the trajectories to reduce the size of the flattened trajectory. The task and style encoders receive the flattened trajectory as input and map it to their respective latent spaces. The latent task and style vectors are then appended and input to the trajectory decoder which outputs the flattened trajectory. We train the networks using the Adam optimizer at a learning rate of 0.0008. Our code can be found here: \url{https://github.com/VT-Collab/PECAN}

Minimizing $\mathcal{L}_{trajectory}$ trains the networks to accurately reconstruct the trajectories in our dataset.
}
However, this still does not guarantee that the representation captured in $Z{\tau}$ aligns with the actual tasks $\mathcal{T}$, or --- along the same lines --- that the representation in $Z_{\theta}$ aligns with the styles $\theta$.
To address this, we propose using a small set of labels $\mathcal{Y}$ for trajectories in the dataset. These labels identify trajectories from \textit{different tasks} that share \textit{similar styles}. In Section~\ref{sec:method-learn}, we will introduce an additional loss function that leverages these labels to ensure that the representation captured in $Z_{\theta}$ aligns with the styles $\theta$. Further, we will show that by minimizing this loss together with $\mathcal{L}_{trajectory}$, we can also ensure that the latent tasks $z_{\tau}$ align with the actual tasks $\tau$.


\subsection{Characteristics of a User-Friendly Canonical Space}\label{sec:method-characteristics}

So far we have discussed how our approach can accurately reconstruct robot trajectories using latent representations of the tasks and styles. However, merely learning latent spaces that can map to the correct behavior is not sufficient, since --- by themselves --- these latent spaces may not be easy for humans to interact with. 
Our goal is to learn a canonical space where the human can intuitively click around the manifold to specify their desired style. 
For instance, given a range of values from $-1$ to $+1$ as shown in \fig{method}, how will the user know which regions of the canonical space correspond to straight or exaggerated trajectories? We now propose ways to structure the robot's learning so that the resulting canonical space is more user-friendly.


One way users can build an understanding of the canonical space is by selecting $z_\theta$ values and then visualizing the resulting robot trajectories across tasks $z_\tau$. 
For example, in \fig{method} the human might click on $z_\theta = +1$ in the learned canonical space, and then observe how the robot arm moves in a straight line to its goal.
But for this interactive approach to be effective, the canonical space needs to be organized such that users can quickly find their desired style by visualizing as few $z_\theta$ values as possible. 
We therefore propose structuring the canonical style space to have the following user-friendly characteristics:
\begin{itemize}
   
    \item \textit{Consistency.} The latent values should result in consistent robot styles $\theta$ across tasks. For example, if $z_{\theta} = +1$ corresponds to a straight line path for the task of placing a cup, the same $z_{\theta}$ should also result in a straight line path for the task of pouring coffee. Therefore, for any $z_{\theta} \in Z_{\theta}$:
    \begin{equation}
        \theta(\phi(z_{\tau}, z_{\theta})) \approx \theta(\phi(z_{\tau}', z_{\theta})) \qquad \forall \; z_{\tau}, z_{\tau}' \in Z_{\tau}
    \label{eq:consistency}
    \end{equation}
    This will allow users to customize the robot's style across all tasks by setting the desired latent style just once.

    \item \textit{Monotonicity.} The styles should vary monotonically as we move from one point in the latent space to another. For example, decreasing the latent style value from $z_{\theta}=+1$ to $z_{\theta} = -1$ should gradually change the robot's style from straight-line paths to increasingly exaggerated arm motions. Therefore, for \reb{one dimensional styles:
    \begin{equation}
        (z_{\theta} - z_{\theta}')(\theta(\phi(z_{\tau}, z_{\theta})) - \theta(\phi(z_{\tau}, z_{\theta}'))) \geq 0
        \qquad \forall \; z_{\theta}, z_{\theta}' \in Z_{\theta}
    \label{eq:monotonous}
    \end{equation}}
    This will enable users to easily find their desired style by interpolating between the extreme ends of the canonical space. 
\end{itemize}


\subsection{Semi-supervised Learning}
\label{sec:method-learn}

We now present our complete training process for learning latent representations of the actual tasks and styles in our dataset, and inducing the user-friendly characteristics --- \textit{consistency} and \textit{monotonicity} --- in the learned canonical space.

\p{Labeling Style Extremes}
We first obtain a small set of labels $\mathcal{Y}$ for trajectories having similar styles but from different tasks. Specifically, we only obtain labels for trajectories that represent the \textit{extremes} of the style range. \reb{For the example in \fig{method}, users may label the most exaggerated trajectories as $y_{1}$ and the least exaggerated arm motions as $y_{2}$.} 
We believe that labeling the extremes is easier than labeling intermediate styles. For instance, users find it difficult to distinguish between slightly different arm motions~\cite{biyik2019asking}. Note that we do not ask users to specify the actual style of a trajectory. We only assume that trajectories with the same label have similar styles, and correspond to one of the extremes of the canonical space.

\p{Style Classifier}
Minimizing the loss $\mathcal{L}_{trajectory}$ introduced in Section~\ref{sec:method-encode} trains the decoder to accurately reconstruct trajectories. Here we include an additional loss to ensure that the canonical space captures the actual styles and is \textit{consistent} across tasks and \textit{monotonic} along each axis. 

We consider each $y_{i} \in \mathcal{Y}$ to be a separate class with one-hot labels $c(y_{i})$, where $\mathcal{Y}$ contains $m$ classes. \reb{For instance, in the above example with labels $\mathcal{Y} = \{y_{1}, y_{2}\}$, we would have $m=2$ classes, e.g., $c(y_{1}) = [1, 0]$ and $c(y_{2}) = [0, 1]$.}
At training time, we pass the labeled trajectories $\xi_{j} \in y_{i}$ through the style encoder $\psi_{\theta}$ to obtain their latent styles $z_{\theta, j}$. We then feed the latent styles into a classifier network $\Delta$ that maps each latent value to a $m$-dimensional vector of class probabilities $p_{j} = [p_{1},  \ldots, p_{m}]$ such that $\sum_{k=1}^{m} p_{j, k} = 1$ for any $p_{j} \in \mathcal{P}$. 
$$\Delta: Z_{\theta} \mapsto \mathcal{P}$$
The classifier consists of a single fully-connected layer followed by a softmax layer. We train the style encoder and classifier to predict the class labels by minimizing the cross-entropy loss:
\begin{equation}
        \mathcal{L}_{ce} = - \sum_{y_{i} \in \mathcal{Y}} \, \sum_{\xi_{j} \in y_{i}} \, 
        \sum_{k=1}^{m} \, c_{k}(y_{i}) \, \log( p_{j, k})
    \label{eq:cross}
\end{equation}
The subscript $k$ refers to the $k$-th index in the $m$-dimensional vectors of class labels and their probabilities. Intuitively, this loss encourages 
trajectories with the same style label $y$ to encode to nearby values within the canonical space, and trajectories with different labels to map to values far from one other in the canonical space.
Particularly, since we only obtain labels for trajectories that represent the extremes of the styles spectrum, the latent style for each extreme will be placed in opposite corners of the canonical space. 

We apply the following theorem from prior work~\cite{graf2021dissecting} to show that when the $\mathcal{L}_{ce}$ loss is minimized, trajectories with the same label are encoded to the same latent value and the value for each label is positioned in a different corner of the latent space.

\p{Theorem} 
Consider a latent space $Z = \{z \in \mathbb{R}^{d} : ||z|| \leq \rho_{Z} \}$ with a radius of $\rho_{Z} > 0$ and a linear classifier with weights $W \in \mathbb{R}^{m \times d}$. Given $N$ latent values $Z = \{z_{1}, \ldots, z_{N}\}$ from this space and a balanced set of labels $Y$, the cross-entropy loss $\mathcal{L}_{ce}$ is bounded as:
\begin{equation}
    \mathcal{L}_{ce}(Z, W; Y) \geq \\ \log \left(1 + (m - 1) \exp \left( - \rho_{Z} \frac{\sqrt{m}}{m - 1} ||W||_{F}\right)  \right)
\end{equation}

This bound is tight if there are points $\zeta_{1}, \ldots, \zeta_{m} \in \mathbb{R}^{d}$ that satisfy the following conditions \cite{graf2021dissecting}:
\begin{enumerate}[itemsep=0.5em]
    \item $\forall n \in [N] : z_{n} = \zeta_{y_{n}}$
    \item $\{\zeta_{y}\}_{y}$ form a $\rho_{\mathcal{Z }}$-sphere-inscribed regular simplex
    \item $\exists \rho_{\mathcal{W}} > 0 : \forall y \in \mathcal{Y} : w_{y} = \frac{\rho_{\mathcal{W}}}{\rho_{Z}} \zeta_{y}$
    \vspace{0.5em}
\end{enumerate}

Condition (1) states that loss $\mathcal{L}_{ce}$ is minimized when latent values $z_{n}$ with the same label $y_{n}$ converge to a common point $\zeta_{y_{n}}$ in the latent space. This means that trajectories with the same style label will be encoded to the same latent style value even if they belong to different tasks.

Conditions (2) and (3) state that the points $\zeta_{1}, \ldots, \zeta_{m}$ and weights corresponding to each class must inscribe a regular simplex in the latent space. This means that the latent values will be positioned at the edges of the latent space, with the points for different labels being equally distant from one another. For instance, consider the $1$D canonical space illustrated in Fig.~\ref{fig:method}. If we have labels for $m=2$ classes representing the style extremes, the latent values for the labeled trajectories will converge to two distinct points $(\zeta_{1}, \zeta_{2})$ --- one for the most exaggerated trajectories and one for the least. A regular simplex inscribed in this canonical space would be a line between end-points $\zeta_{1} = -1$ and $\zeta_{2} = +1$. Therefore, when $\mathcal{L}_{ce}$ is minimized, the latent values for the style extremes will be pushed to the opposite ends of the canonical space.

We take advantage of these conditions to learn a canonical space that captures the actual styles and exhibits the desired user-friendly characteristics as follows:

\p{Combined Loss} We simultaneously train all networks by minimizing the combined loss $\mathcal{L}$.
\begin{equation}
    \mathcal{L} = \mathcal{L}_{trajectory} + \mathcal{L}_{ce} \label{eq:combined}
\end{equation}
First, we examine how minimizing $\mathcal{L}$ allows us to represent the actual tasks and styles in the data using their respective latent spaces. According to condition (1), minimizing the $\mathcal{L}_{ce}$ loss causes all trajectories with the same label to be encoded to the same latent value. For example, consider a label with two trajectories, $y_{i} = \{\xi_{1}, \xi_{2}\}$. When $\mathcal{L}_{ce}$ is minimized, the style encoder $\psi_{\theta}$ will map both trajectories to the same latent style, i.e., $\psi_{\theta}(\xi_{1}) = \psi_{\theta}(\xi_{2}) = z_{\theta, i}$. To simultaneously minimize $\mathcal{L}_{trajectory}$, the decoder $\phi$ must reconstruct the trajectories from this same latent style value. Recall that we assume each label has trajectories with similar styles but from different tasks, meaning $\xi_{1} \neq \xi_{2}$. To output two different trajectories given the same latent style as input, i.e., $\phi(\psi_{\tau}(\xi_{1}), z_{\theta, i}) \neq \phi(\psi_{\tau}(\xi_{2}), z_{\theta, i})$,  the decoder will require the trajectories to have different latent task values. Therefore, the task encoder must learn to map these trajectories to distinct values in the latent task space, i.e., $\psi_{\tau}(\xi_{1}) \neq \psi_{\tau}(\xi_{2})$. In this way, we can train the task and style encoders to embed the actual tasks and styles in their respective latent spaces.

Next, we explore how condition (1) enables the style encoder to construct a \textit{consistent} canonical space. Following the previous example, we see that minimizing $\mathcal{L}_{ce}$ trains the style encoder to map trajectories from different tasks to the same latent value. Since these trajectories belong to the same label, they have the same actual style. 
This results in a \textit{consistent} canonical space where a given latent value corresponds to trajectories with similar styles across different tasks.

Lastly, according to condition (2), minimizing $\mathcal{L}_{ce}$ places the latent values for the extreme styles at the opposite ends of the canonical space. In practice, we find that training the style encoder to minimize both $\mathcal{L}_{trajectory}$ and $\mathcal{L}_{ce}$ causes the latent values of trajectories with intermediate styles to be placed \textit{monotonically} between the extremes.
\reb{We use equal weights for both losses when optimizing the combined loss in our experiments.}

\p{Summary}
At training time, our architecture leverages a dataset of user demonstrations $\mathcal{D}$ and a small set of labels $\mathcal{Y}$ to learn a latent task space $Z_{\tau}$ and a canonical space of styles $Z_{\theta}$. We structure the canonical space to be consistent and monotonic by optimizing the combined loss in \eq{combined}. \reb{At runtime, the user can select any point $z_{\theta}$ inside the learned canonical space.} The decoder $\phi$ then takes this latent style as input and reconstructs the corresponding robot trajectory for each $z_{\tau} \in Z_{\tau}$.

In the following sections we will experimentally demonstrate the ability of our proposed architecture to learn distinct latent spaces for tasks and styles. We will also evaluate the accuracy of the trajectories generated from these latent representations, and assess whether the learned canonical space maintains our desired, user-friendly characteristics.

\section{Simulation Experiments}
\label{sec:simulations}

\begin{figure*}[t]
    \begin{center}
        \includegraphics[width=1\textwidth]{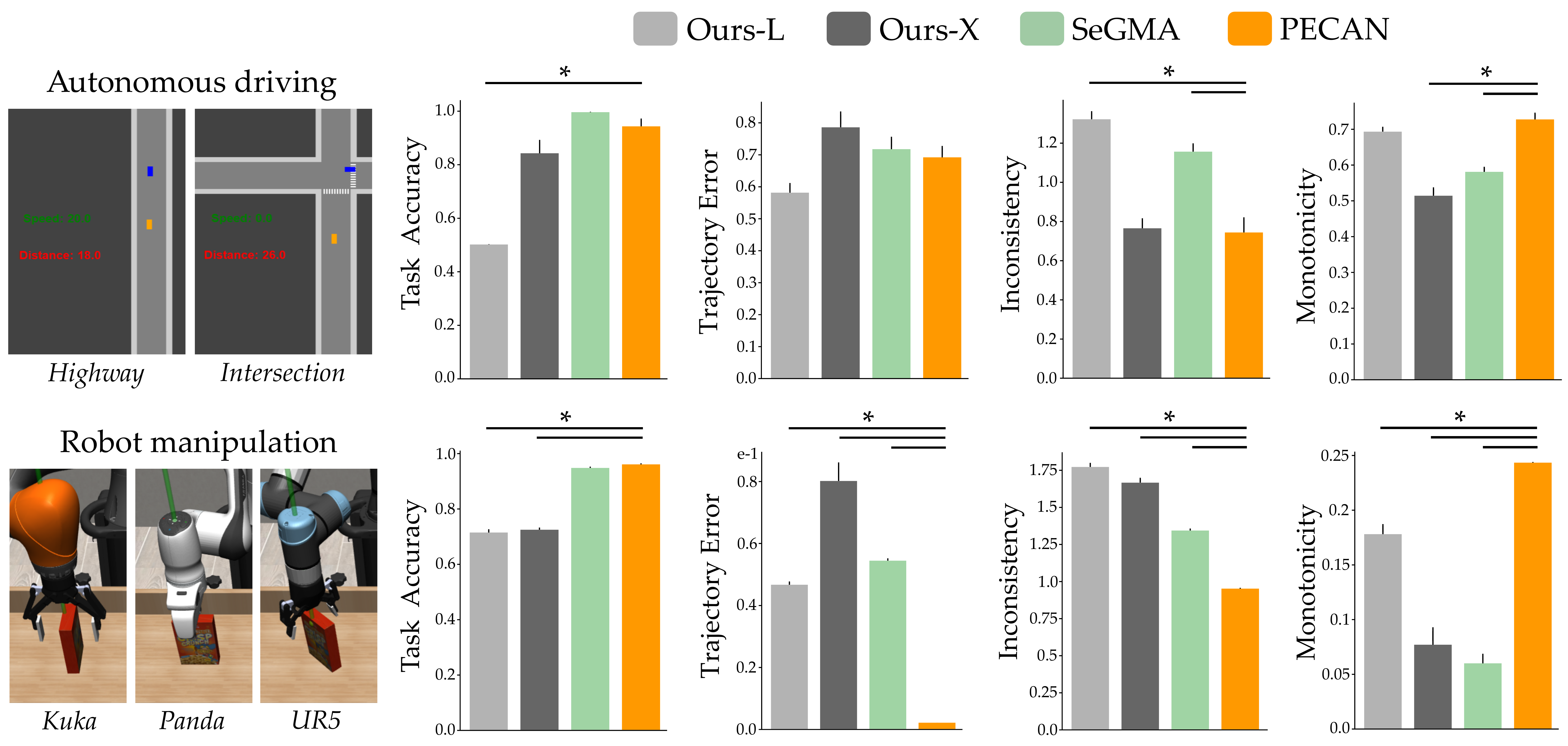}
        \caption{Simulation results for autonomous driving (Top row) and robot manipulation (Bottom row) environments. We compare our proposed approach, PECAN, to a state-of-the-art baseline, SeGMA, and ablations of our approach, Ours-L and Ours-X. While SeGMA uses task labels, PECAN uses labels for trajectories with similar styles (specifically the style extremes). Both ablations use the same architecture as PECAN, however, Ours-L does not train with any labels, whereas Ours-X uses labels for trajectories with intermediate styles (instead of the style extremes). In both environments, PECAN achieves comparable \textit{Task Accuracy} to SeGMA. Although PECAN has significantly lower Trajectory Error in the robot environment, its performance is similar to the baselines in the driving environment. The main advantage of PECAN over the baseline methods is that the canonical spaces learned by our approach are more consistent and monotonic (i.e., more user friendly). An asterisk (*) denotes statistical significance \reb{and the error bars indicate standard error.}}
        \label{fig:simulation}
    \end{center}
\end{figure*}

Here we perform controlled simulations to analyze the contributions of each component of PECAN.
We compare the performance of our proposed approach to a state-of-the-art baseline for learning latent style representations~\cite{smieja2020segma} and ablations of our method in two environments: autonomous driving and robot manipulation (see \fig{simulation}).

\p{Environments} In the first environment we personalize the driving style of an autonomous car across two tasks, \textbf{Highway} and \textbf{Intersection}. In \textit{Highway} the autonomous car follows another car on a highway. In \textit{Intersection} the autonomous car waits for another car to pass before safely crossing an intersection. 
\reb{
The states include the 2D positions of both cars, $s = [x_{auto}, y_{auto}, x_{other}, y_{other}]$, and the actions are the autonomous car's velocity, $a = [\Delta x_{auto}, \Delta y_{auto}].$
}
In both tasks, we consider 2D styles $\theta = [\theta_{1}, \theta_{2}]$ that define how aggressively or defensively the car drives. Here $\theta_{1}$ corresponds to the maximum speed the car achieves in an empty section of the road, and $\theta_{2}$ represents the minimum distance that the car keeps from other vehicles on the road. For example, some users may prefer a high speed of $\theta_{1} = 100$ km/h until their car is within $\theta_{2} = 30$ feet of the next car. Other users may prefer a slow speed of $\theta_{1} = 40$ km/h but get as close as $\theta_{2} = 10$ feet of the next car. We implement these tasks using the CARLO simulator~\cite{cao2020reinforcement}.

In the robot environment we move away from the conventional meaning of tasks and styles to showcase the versatility of our approach. 
We consider three different robot platforms as the tasks: a \textbf{Kuka}, \textbf{Panda}, and \textbf{UR5}. In each task (i.e., for each type of robot) the goal is to transfer a cereal box from one bin to another. 
\reb{
The states are the joint positions and gripper configurations of the respective robot platforms and the actions are their joint and gripper velocities. Since the arms and grippers of each robot have different degrees of freedom (DoFs) --- the \textit{Kuka} has a $7$ DoF arm and a $6$ DoF gripper, the \textit{Panda} also has a $7$ DoF arm but a $2$ DoF gripper, while the \textit{UR5} has a $6$ DoF arm and $6$ DoF gripper --- we append zeros to the states and actions of the \textit{Panda} and \textit{UR5} robots to ensure that all trajectories have the same number of dimensions.
}
The styles are three dimensional $\theta = [\theta_{1}, \theta_{2},  \theta_{3}]$, and represent variations in environment. The variable $\theta_{1}$ is the orientation of the cereal box, while $\theta_{2}$ and $\theta_{3}$ mark the position of the target bin along the $x$ and $z$ axis respectively. We implement this environment using Robosuite~\cite{robosuite2020}.

\p{Baselines} We compare \textbf{PECAN} to the following methods:
\begin{itemize}
    \item \textbf{Ours-L}: An ablation of our approach that does not use any labeled data and only trains using the $\mathcal{L}_{trajectory}$ loss. Since it does use labels for trajectories with similar styles, we expect this approach to learn a canonical space that is not consistent across tasks, similar to prior work~\cite{osa2020goal}.
    \item \textbf{Ours-X}: An ablation of our approach that uses labels for trajectories with intermediate styles, instead of the style extremes. We expect such an approach to learn a canonical space that is consistent but not monotonous.
    \item \textbf{SeGMA}: A state-of-the-art approach for learning latent styles across multiple classes~\cite{smieja2020segma}. Instead of learning two separate task and style spaces, this method learns one combined latent space where the classes (i.e., tasks) are represented as Gaussians and their variance captures the styles. A latent style $z_{s}$ can be transferred from one task centered at $\mu_{s}$ to another task centered at $\mu_{t}$ by:
    $$z_{t} = z_{s} + (\mu_{t} - \mu_{s})$$
    This approach uses task labels instead of labels for trajectories with similar styles. Therefore we do not expect the latent styles to be consistent across tasks.
\end{itemize}

\reb{
We assume the dimensionality of the styles is known and model the latent style space for all methods to have the same number of dimensions as the true styles. We use a $d_{\theta} = 2$ dimensional latent space for the driving environment and a $d_{\theta} = 3$ dimensional space for the robot environment. Likewise, we set the number of latent tasks to match the number of tasks in each environment.
}

\p{Training} In each environment we obtain a set of trajectories $\Xi$, where every trajectory $\xi \in \Xi$ corresponds to a different task and style $(\tau, \theta)$. In the driving environment, the trajectories are generated by simulated humans with different styles, while in the robot environment, the trajectories are teleoperated by an expert user. Next, we sample a small set of demonstrations $\mathcal{D}$ from the full set of trajectories $\Xi$ to train the methods. In the driving environment, we create a training dataset of $16$ demonstrations from a set of $352$ trajectories such that $8$ demonstrations belong to \textit{Highway} and $8$ belong to \textit{Intersection}. Four trajectories in each task correspond to the extreme styles --- [high $\theta_{1}$, high $\theta_{2}$], [high $\theta_{1}$, low $\theta_{2}$], [low $\theta_{1}$, high $\theta_{2}$], and [low $\theta_{1}$, low $\theta_{2}$] ---  while the other four are randomly sampled. In the robot environment, we sample a training dataset of $27$ demonstrations from a set of $60$ trajectories. The data is balanced across the three tasks. Of the $9$ demonstrations in a task, $8$ represent the style extremes and $1$ is randomly sampled. 

For \textbf{Ours-X}, all trajectories in a task are randomly sampled.
\reb{
We train all methods using the same number of demonstrations $\mathcal{D}$ and labels $\mathcal{Y}$, except for \textbf{Ours-L}, which does not use the labels and only learns to reconstruct the demonstrations.}
\textbf{SeGMA} requires labels for trajectories that belong to the same task. In contrast, \textbf{Ours-X} and \textbf{PECAN} do not require any task labels and only use labels for trajectories with similar styles.
\reb{These labels are provided by the expert user.}

\p{Testing} We test the performance of each method using the entire set of trajectories $\Xi$. Specifically, we measure the accuracy of encoding the tasks (\textit{Task Accuracy}), the error in reconstructing the trajectories (\textit{Trajectory Error}), and the \textit{Consistency} and \textit{Monotonicity} of the latent style space.

\textit{Task Accuracy} is $1$ if trajectories that belong to the same task are encoded to the same value in the latent task space, with distinct latent values for trajectories in different tasks. If all trajectories are mapped to the same latent task, the \textit{Task Accuracy} is $1 / |\mathcal{T}|$. \textit{Trajectory Error} is the mean squared error between the original trajectory and the trajectory reconstructed from the latent style space. Next, as a proxy for measuring the \textit{Consistency} of the canonical space, we measure its \textit{Inconsistency} by computing the difference between the latent style values of trajectories that have similar styles but in different tasks. Inconsistency is defined as: 
\begin{equation}
    \mathbb{E}_{\xi_{1}, \xi_{2} \in \Xi} \; ||\psi_{\theta}(\xi_{1}) - \psi_{\theta}(\xi_{2})|| \quad \text{if} \;\; \theta(\xi_{1}) = \theta(\xi_{2}) \;\; \text{and} \;\;  \tau(\xi_{1}) \neq \tau(\xi_{2})
\end{equation} 
Here we do not compute \textit{Consistency} directly using \eq{consistency} because it requires access to a function $\theta(\cdot)$ that maps reconstructed trajectories to their actual style values. In our simulations, we only know the actual tasks and styles for the trajectories in the dataset $\Xi$, \reb{but} not for their reconstructions.
Lastly, we measure the \textit{Monotonicity} of the canonical spaces by computing the correlation between the latent values and the actual styles of trajectories in $\Xi$. In a monotonous space, 
\reb{
the difference in latent values of trajectories $\Delta Z_{\theta}$ should be rank correlated to the difference in their styles $\Delta \theta$. We measure this by calculating the Spearman's rank correlation coefficient~\cite{zar2005spearman} between the ranks of $\Delta Z_{\theta}$ and $\Delta \theta$.
\begin{equation}
\begin{aligned}
    \Delta\theta = [||\theta(\xi_{1}) - \theta(\xi_{2})|| \;\; | \;\; \forall \, \xi_{1}, \xi_{2} \in \Xi ] \\
    \Delta Z_{\theta} = [||\psi_{\theta}(\xi_{1}) - \psi_{\theta}(\xi_{2})|| \;\; | \;\; \forall \, \xi_{1}, \xi_{2} \in \Xi ]
\end{aligned}
\end{equation}}
A coefficient of $1$ or $-1$ indicates perfect correlation, while $0$ means that the styles and their latent values are uncorrelated. We take the absolute value of this coefficient as the \textit{Monotonicity} of the canonical space.

\p{Results}
Our results are displayed in \fig{simulation}. We calculated these results over $20$ training runs, each starting with randomly initialized network weights. 

We performed one-way ANOVA tests and found that the choice of method had a significant effect on \textit{Task Accuracy} in the autonomous driving ($F(3, 76)=58.7$, $p<0.01$) and robot manipulation ($F(3, 76)=313.1$, $p<0.01$) environments. In both environments, \textbf{SeGMA} achieved a high \textit{Task Accuracy}, while \textbf{Ours-L} achieved the lowest. This is likely because \textbf{SeGMA} is trained with labels for the tasks, whereas \textbf{Ours-L} operates without any labels. \textbf{PECAN}, on the other hand, does not use task labels like \textbf{SeGMA}. Yet it achieved a comparable \textit{Task Accuracy} by leveraging labels for trajectories with similar styles. \reb{Pairwise comparisons using Tukey's HSD post-hoc test} indicated a statistically significant difference ($p < 0.01$) between the \textit{Task Accuracy} of \textbf{PECAN} and \textbf{Ours-L} in both environments. On the other hand, there was no significant difference in the \textit{Task Accuracy} of \textbf{PECAN} and \textbf{SeGMA} in the autonomous driving ($p=0.57$) and robot manipulation ($p=0.64$) environments. \reb{The p-values have been adjusted for multiple pairwise comparisons.}

\textbf{Our-X} also attains high \textit{Task Accuracy} in the driving environment by leveraging the style labels similarly to \textbf{PECAN} ($p=0.07$). However, in the robot environment, \textbf{Ours-X} has a significantly lower \textit{Task Accuracy} ($p<0.01$) due to a high error in its reconstructed trajectories. These findings demonstrate that to learn the correct task representation by only using labels for trajectories with similar styles, it is crucial to optimize both the reconstruction loss and the cross-entropy loss, as we theoretically suggested in Section \ref{sec:method-learn}. 

\reb{
Although \textbf{Ours-L} struggled to encode the tasks, it had the lowest error in reconstructing the trajectories in the driving environment. Unlike other methods that needed to balance trajectory reconstruction with shaping the canonical space, \textbf{Ours-L} solely focused on minimizing the trajectory loss. A one-way ANOVA test revealed a significant effect of the choice of method on \textit{Trajectory Error} ($F(3, 76)=4.7$, $p<0.05$) in the driving environment. However, despite of achieving the lowest trajectory error, a Tukey's HSD post-hoc test did not indicate a significant difference ($p=0.2$) between \textbf{Ours-L} and \textbf{PECAN}. On the contrary, we found that \textbf{Ours-L} had significantly higher \textit{Trajectory Error} ($p<0.01$) than \textbf{PECAN} in the robot environment. We realized that \textbf{Ours-L} tended to overfit to the training demonstrations in this environment leading to a poor test performance.

Finally}, we observed that the canonical spaces learned by \textbf{PECAN} are more consistent and monotonous than those learned by any of the baselines. One-way ANOVA tests revealed that the choice of method had a significant effect on the \textit{Consistency} ($F(3, 76)=27.5$, $p<0.01$) and \textit{Monotonicity} ($F(3, 76)=30.3$, $p<0.01$) of the canonical space in the driving environment. \reb{A Tukey's HSD post-hoc test} indicated that the spaces learned by \textbf{Ours-L} show comparable \textit{Monotonicity} ($p=0.53$) to \textbf{PECAN} but lack \textit{Consistency} ($p<0.01$) because of not using any labels. In contrast, \textbf{Ours-X} manages to learn a consistent latent space by using the style labels similar to \textbf{PECAN} ($p=0.99$) in the driving environment. However, it lacks \textit{Monotonicity} ($p<0.01$) because it obtains labels for the intermediate styles rather than the style extremes. \textbf{SeGMA} does not leverage any style labels and thus has a lower consistency ($p<0.01$) and monotonicity ($p<0.01$) than \textbf{PECAN}.

\p{Takeaways}
These results demonstrate that PECAN can successfully learn the task and style encodings by only using labels for trajectories with similar styles. While SeGMA achieves similar accuracy in encoding the tasks and reconstructing trajectories by using task labels, unlike PECAN, it does not learn a consistent and monotonic canonical space. We hypothesize that these characteristics make the canonical space intuitive for users to understand, enabling them to easily find a latent value corresponding to their preferred style.

Our ablations highlight that each component of PECAN is critical for constructing a user-friendly canonical space. The style labels ensure that PECAN learns a consistent canonical space and obtaining these labels for trajectories with extreme styles helps in making the space monotonic. Conversely, the canonical space learned by Ours-L is inconsistent because it does not use any labels, while the canonical space learned by Ours-X is consistent but not monotonic because it uses labels for intermediate styles as opposed to the style extremes.
\section{User study}

Our simulated experiments indicate that our proposed approach learns a consistent and monotonic space of styles. In this section, we will investigate whether these characteristics actually make the canonical spaces user-friendly, and whether users are able to leverage these spaces to directly personalize robot behaviors.
 
We conducted two in-person user studies to evaluate the effectiveness of PECAN with real users. In the first study, we compare two direct approaches for personalizing the trajectory of a robot arm using learned canonical spaces. Specifically, we test whether the consistent and monotonic spaces learned by PECAN are more \textit{intuitive} to users than the spaces learned by the state-of-the-art baseline, SeGMA. In the second study, we address the overarching question of how to best personalize robot styles: through \textit{direct} selection in a style space, or via \textit{indirect} methods that learn from user feedback. We compare our direct approach, PECAN, with a standard indirect method from prior work~\cite{sadigh2017active}. We test these approaches in the context of customizing the driving style of an autonomous car and assess the pros and cons of each method.


\subsection{Learning User-Friendly Canonical Spaces}

In our first user study, we tested if structuring the canonical spaces to be consistent and monotonic makes them more user-friendly for personalizing robot styles. A user-friendly space should be intuitive and easy to use, enabling users to quickly find their desired style. We compared two approaches for learning a canonical space of robot styles: PECAN and SeGMA~\cite{smieja2020segma}. Our simulations in Section~\ref{sec:simulations} showed that both approaches effectively learned latent representations of the tasks and styles from robot trajectories. However, the spaces learned by PECAN were more consistent and monotonic as compared to those learned by SeGMA. Therefore, we hypothesized that users would find PECAN to be more intuitive and easier to use than SeGMA for personalizing robot styles.

\begin{figure*}[t]
    \begin{center}
        \includegraphics[width=1\textwidth]{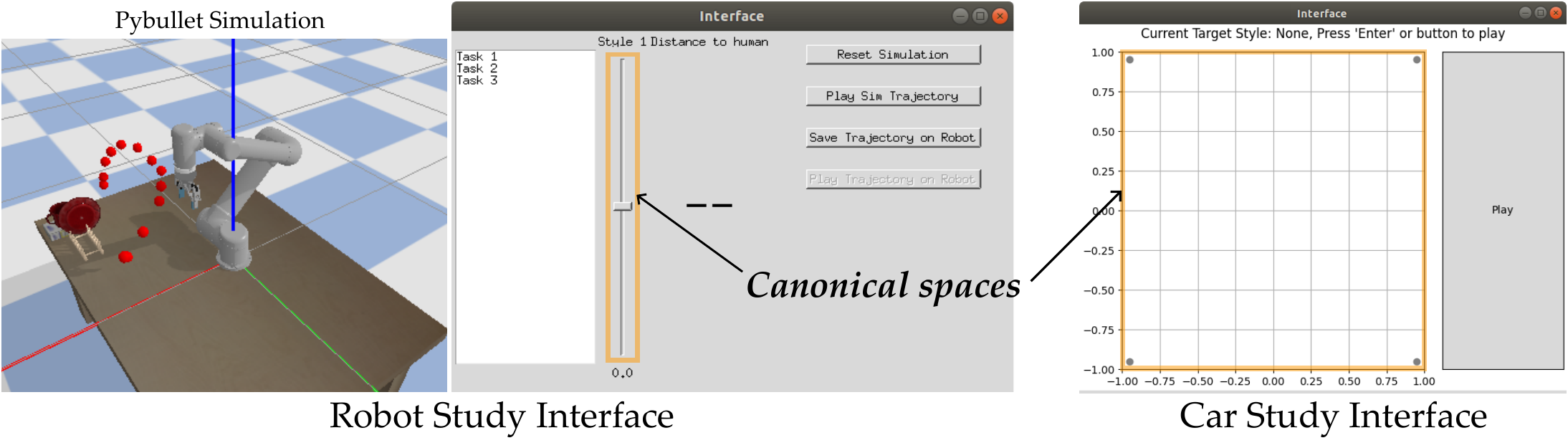}
        \vspace{-2ex}
        \caption{Interfaces for personalizing the behavior of the robot (Left) and the autonomous car (Right) in our user studies. In the robot study, the canonical space was a 1D line which represented the distance that the robot maintains from the user. Users personalized the style of the robot's trajectory by moving the slider along the line. For tasks 1 and 2, users could visualize the robot's trajectory in a Pybullet simulation before executing it on the robot in the real world. In the car study, the canonical space was a 2D square which captured the maximum speed of the autonomous car and the minimum distance it maintains from other cars on the road. Users personalized the car's driving style by selecting different points inside the square. Since the driving environment was entirely in simulation, there was no need to visualize the car's trajectory separately before execution.
        \fig{simulation} shows examples of the simulated car in the Highway and Intersection tasks.}
        \label{fig:interfaces}
    \end{center}
\end{figure*}

\p{Experimental Setup}
Participants in this study interacted with a $6$-DoF UR5 robot arm in three manipulation tasks: (i) handing over a plate to the user (\textbf{Handover Plate}), (ii) placing a cup in front of the user (\textbf{Place Cup}), and (iii) pouring coffee into the cup (\textbf{Pour Coffee}). 
\reb{
The states and actions for each task were the robot arm's joint positions and velocities, respectively.

T}he robot's style was defined by the distance it maintained from the user. At one extreme, the robot could follow a straight-line path that comes very close to the user, while at the other extreme, it could take a curved path that stays as far from the user as possible. 
To personalize the robot's trajectory, each user interacted with an interface containing a canonical space of the robot's styles. The interface featured a task selection menu, a slider for choosing the latent style (as shown in Fig.~\ref{fig:interfaces}-Left), and buttons to execute the trajectory corresponding to the chosen style in a Pybullet simulation and on the real robot. We pre-programmed the robot to pick up the objects for each task  (i.e., plate, cup, or kettle) from their initial positions. \reb{Users first chose a task from the menu, which mapped to a latent task vector, then moved the slider to select the latent style.} The robot then performed the task according to the style selected by the user. See videos here: \url{https://youtu.be/wRJpyr23PKI}

Participants performed the three tasks in the order given above. In the first two tasks, \textit{Handover Plate} and \textit{Place Cup}, they could preview the robot's trajectory in the Pybullet simulation before executing it in the real world. However, they did not have this option in the third task of \textit{Pour Coffee} and had to find their target style based on their experience of using the interface in the first two tasks. We did this to determine if users could build an accurate understanding of the canonical space, enabling them to effectively transfer it to new tasks. Therefore, we treated the first two tasks as familiarization tasks and evaluated the user performances in the \textit{Pour Coffee} task. \reb{We chose the \textit{Pour Coffee} task for user evaluation because it was the most visually distinct among the three tasks, making it challenging for users to find their target style without visualization. We informed users that they would not be allowed to visualize the robot's trajectory in simulation for the \textit{Pour Coffee} task and must use the first two tasks to learn how the latent values mapped to the robot's style.} We anticipated that a consistent and monotonic canonical space would be easier for users to learn and apply across new tasks without the need to visualize the robot's behavior.

\p{Independent Variables} 
We compared our proposed approach (\textbf{PECAN}) to a state-of-the-art baseline for generating latent style representations (\textbf{SeGMA}). 
\reb{To create the training dataset, we first provided three demonstrations each in the \textit{Handover Plate} and \textit{Place Cup} tasks. Two of the demonstrations in each task represented the extreme styles --- trajectories with the maximum and minimum distance to the user. These $6$ demonstrations were common across all participants. In addition to the demonstrations provided by experimenters, we asked each participant to provide $2$ demonstrations in the \textit{Pour Coffee} task, one for each of the extreme styles. In total, we trained custom PECAN and SeGMA models for each participant using a dataset of 8 demonstrations along with labels for the trajectories representing the extreme styles. For PECAN, we asked participants to assign the same label to trajectories with similar styles across the tasks. On the other hand, an expert provided the same label to trajectories in the same task for SeGMA.
We modeled both approaches to have a $d_{\theta} = 1$ dimensional latent style space and $d_{\tau} = 3$ latent tasks.
}


\begingroup
\renewcommand{\arraystretch}{1.1}
\begin{table}[t!]
\centering
\caption{Survey questions (Likert scales with $7$-option response format). \reb{We grouped the questions for the robot study into five scales: \textit{Easy}, \textit{Intuitive}, \textit{Accurate}, \textit{No Visuals}, and \textit{Prefer}. We included additional questions for the \textit{Learn} scale in the autonomous car study. We tested the reliability of each scale using Cronbach's $\alpha$. The reliability scores presented for the first five scales are based on the responses recorded in the robot study. The reliability score for the Learn scale is based on user responses in the autonomous car study.}}
\label{tab:questions}
\resizebox{\linewidth}{!}{%
\begin{tabular}{l c}
\hline
Questionnaire item & Reliability\\
\hline
\textbf{Easy} \\
- It was easy to personalize the robot trajectory using this interface.  & 0.89\\
- It was challenging to personalize the robot trajectory using this interface.\\ 
\hline
\textbf{Intuitive}\\
- I was able to understand how moving the slider changed the robot trajectory. \\
- It was difficult to understand how the robot trajectory would change by moving the slider. & 0.94\\
- The interface was intuitive to use for personalizing the robot trajectory.\\ 
- I did not find the interface intuitive for personalizing the robot trajectory.\\ 
\hline
\textbf{Accurate}\\
- In the end, I was able to accurately personalize the robot trajectory. & 0.88\\ 
- In the end, I was unable to personalize the robot trajectory accurately.\\ 
\hline
\textbf{Easy (No Visuals)} \\
- It was easy to personalize the robot trajectory in the third task based on the first two tasks.  & 0.93\\
- It was challenging to personalize the robot trajectory in the third task based on the first two tasks.\\ 
\hline
\textbf{Prefer}\\
- Overall, I would prefer to use this interface to personalize the robot trajectory. & 0.85\\ 
- Overall, I would not like to use this interface to personalize the robot trajectory.\\ 
\hline
\hline
\reb{
\textbf{Learn}}\\
\reb{- I needed fewer tries to personalize the robot as I gained more experience with the interface.} & \reb{0.81}\\ 
\reb{- I needed more tries to personalize the robot as I gained more experience with the interface.}\\ 
\hline
\end{tabular}%
}
\end{table}
\endgroup

\p{Participants and Procedure} 
We recruited $14$ participants ($3$ female, average age $28 \pm 5$ years) from the Virginia Tech community. Participants gave informed written consent under IRB \#23-1237. At the start of the experiment, \reb{we informed the participants that the robot's style refers to the distance it maintains from their body and asked them to} kinesthetically guide the robot arm to demonstrate trajectories for the extreme styles in the \textit{Pour Coffee} task. We then trained both methods on the user demonstrations along with the six previously collected demonstrations. Users interacted with the robot in all three tasks --- once with each method. They completed the tasks in a fixed order but the ordering of the methods was counterbalanced: half of the users started with PECAN and the other half started with SeGMA.

In the \textit{Handover Plate} and \textit{Place Cup} tasks, users personalized the robot's trajectory once to match distinct target styles. Then, in the \textit{Pour Coffee} task, users personalized the robot's motion twice to achieve two additional target styles. We randomly sampled the four target styles for each user. \reb{These styles were different from the demonstrations in the training data and could correspond to any point in the canonical space.} For each style, users had $3$ attempts to perform the task on the real robot while ensuring that its trajectory stayed within a small tolerance of the desired distance. To help users gauge the actual style of the robot's trajectory, we displayed its maximum distance from the user on the interface.


\p{Dependent Variables} 
For each task we recorded the number of attempts that users took to achieve the target style on the real robot (\textbf{Real Attempts}). 
We also recorded the number of times users visualized the latent styles in simulation before executing them in the real world (\textbf{Sim Attempts}) for the \textit{Handover Plate} and \textit{Place Cup} tasks.
A higher number of attempts indicated that it was difficult for users to identify their desired styles in the learned canonical space. Specifically, in the \textit{Pour Coffee} task, where users did not have the option to simulate the styles, a higher number of attempts indicated that the interface was not intuitive and consistent across tasks. We also measured the \reb{absolute} error in the distance (\textbf{Style Error}) and final position (\textbf{Task Error}) of the trajectories executed by users in their last attempt for each target style.

After working with each interface users answered a 7-point Likert scale survey (see Table~\ref{tab:questions}). This survey measured their subjective responses along five scales: the \textit{Intuitiveness} of the interface, how \textit{Easy} it was to personalize the robot's style with that interface, how easy it was to personalize the robot's style without visual information (\textit{No Visuals}), the \textit{Accuracy} of the reconstructed trajectories, and if they \textit{Preferred} using that interface. Users also detailed their experience after using each interface in an open-ended response.


\p{Hypothesis}
We had the following hypotheses:

\begin{quote}

\p{H1} \textit{Users will find interfaces that use PECAN to be easier and more intuitive than those that use SeGMA for personalizing the robot trajectory.}

\p{H2} \textit{Users will subjectively prefer using canonical spaces learned by PECAN over those learned by SeGMA.}

\end{quote}

\begin{figure*}[t]
    \begin{center}
        \includegraphics[width=1\textwidth]{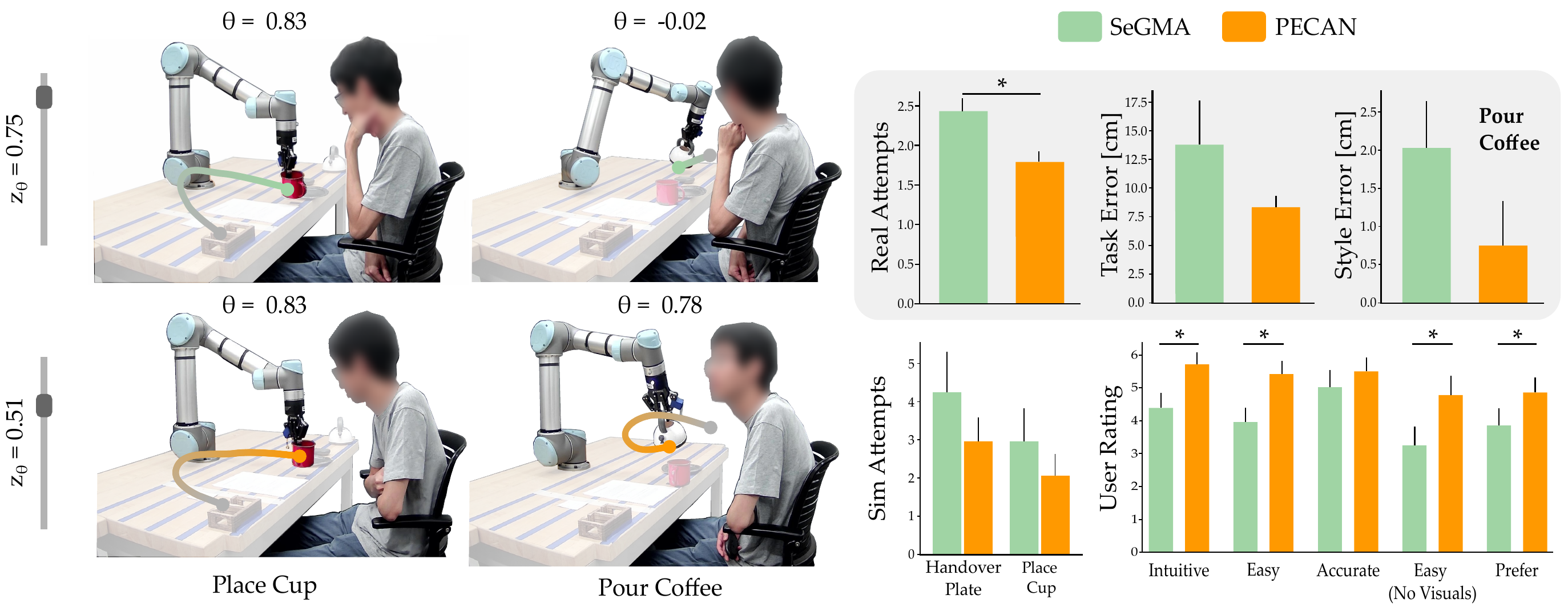}
        \caption{Objective and subjective results from the robot user study. (Left) User applies the same latent value $z_{\theta}$ from \textit{Place Cup} to \textit{Pour Coffee} expecting a similar style across both tasks. PECAN produces trajectories with similar distances $\theta$ for \textit{Place Cup} and \textit{Pour Coffee}, but SeGMA generates a straight line trajectory for \textit{Pour Coffee}. (Top Right) When using PECAN participants had lower \textit{task error}, \textit{style error}, and fewer \textit{real attempts} (t(13) = -3.12, p < 0.01) performing new tasks without visual information. (Bottom Right) Subjectively participants \textit{prefer} (t(13) = 2.55, p < 0.05) working with PECAN compared to SeGMA, as they found PECAN more \textit{intuitive} (t(13) = 3.35, p < 0.01), and \textit{easy} (t(13) = 2.68, p < 0.05) especially without visual information (t(13) = 2.76, p < 0.05). An asterisk (*) denotes statistical significance \reb{and the error bars indicate standard error.}}
        \label{fig:robotstudy}
    \end{center}
\end{figure*}

\p{Results}
Our results are presented in Fig.~\ref{fig:robotstudy}. In the first two tasks users required a similar number of real attempts for both methods. This was because they could spend ample time refining their desired style in simulation before executing it on the real robot. Therefore, we tested our first hypothesis by comparing the performance of the two methods in the \textit{Pour Coffee} task.
A two-tailed paired t-test showed a significant difference in the number of real attempts ($p<0.01$) with PECAN and SeGMA. This suggests that PECAN is more intuitive and consistent than SeGMA, making it easier for users to find their target style. Subjectively, users reported that it was easier to personalize the robot with PECAN than SeGMA, especially in the third task where they had no visual information ($p<0.05$). Users also reported that they found PECAN to be more intuitive than the baseline. A two-tailed paired t-test showed a significant difference in the combined ratings of the \textit{Easy} ($p < 0.05$) and \textit{Intuitive} ($p<0.05$) scales for PECAN and SeGMA. This result supported hypothesis \textbf{H1}.

In total, $11$ out of $14$ users stated that they preferred working with interfaces trained using PECAN, giving it a significantly higher rating than SeGMA on the \textit{Prefer} scale ($p <0.05$). This result supported \textbf{H2}.

\p{Takeaways}
Overall, these results demonstrate that the consistent and monotonous spaces learned by our approach are more intuitive for users, making it easy for them to personalize the robot, especially when they cannot simulate the robot's motion before executing it in the real world. In their open-ended response, five users stated that the slider (i.e., canonical space) for the baseline approach, SeGMA, was inconsistent across tasks. For example, one user wrote that ``\textit{this one [SeGMA] appeared to change for each task, which made it hard to get to understand the slider}''. Therefore, users needed more attempts to achieve their target style in the \textit{Pour Coffee} task with SeGMA as compared to PECAN. We also observed that users achieved a lower task and style error in the \textit{Pour Coffee} task when using PECAN, although the difference was not statistically significant. 

\subsection{Direct vs. Indirect Personalization}\label{sec:indirect}

In the second user study we determined the pros and cons of \textit{directly} selecting the robot's style compared to \textit{indirect} methods that estimate the style from user feedback (e.g., ranking robot trajectories).
\reb{Here we shifted to the driving environment which is a standard benchmark for inferring user preferences and enables us to test the personalization of more complex 2D styles, as opposed to the simpler 1D style in the robot manipulation task.}
Participants were presented with two fundamentally different approaches for modifying an autonomous car's driving style --- our direct approach, PECAN, and an established approach for indirectly learning from human preferences~\cite{sadigh2017active}.

\p{Independent Variables} Specifically, we compared \textbf{PECAN} to an active preference-based learning approach, which we refer to as \textbf{APReL}. We implemented this approach based on the code provided in~\cite{biyik2022APReL}. 
At each step, APReL presented two trajectories to the users, each representing a different style, and asked them to choose their preferred trajectory. Based on their choices, APReL inferred the individual user styles. It strategically selected these trajectories to maximize the information it gained about the user's style from their choice. For example, showing trajectories with distinctly different speeds, such as one high-speed and one low-speed, is more informative than showing two trajectories with similar speeds.
In our implementation, APReL selected the trajectories from a discrete space of $176$ uniformly sampled styles. We trained PECAN with $24$ demonstrations: $8$ demonstrations corresponded to the extreme styles and $16$ were sampled randomly, as explained in Section~\ref{sec:simulations}.
\reb{We used a $d_{\theta}=2$ dimensional latent style space for PECAN and, correspondingly, $2$ dimensional features for APReL.} It is important to note that, unlike our approach, APReL has direct access to the features that parameterize the driving style of the autonomous car. Therefore, we might expect this baseline to outperform our approach because it knows the actual styles, while our approach must learn these styles from user demonstrations.

\p{Experimental Setup} The driving simulation in this study was the same as in Section~\ref{sec:simulations}. We had two tasks, \textbf{Highway} and \textbf{Intersection}, and 2D styles that depended on the speed of the autonomous car and the minimum distance that it maintains from other vehicles. For PECAN, users selected their preferred style by clicking on a point in a 2D canonical space as shown in \fig{interfaces}-Right. In contrast, APReL showed simulations of two car trajectories and asked users to select the trajectory that best matched their preferred style.

\p{Participants and Procedure} We recruited $10$ participants ($2$ female and $1$ undisclosed, average age $27 \pm 5$ years) from the Virginia Tech community. None of these participants took part in our first user study. Participants gave informed written consent under IRB \#23-1237. 

We asked each user to personalize the driving style of the autonomous car to a randomly sampled style (i.e., car speed and following distance) across both tasks. Users customized the car's style using both direct (PECAN) and indirect (APReL) approaches. We counterbalanced the ordering of these approaches.
When using PECAN, users clicked on points in the canonical space and visualized the car behavior until they found their target style. Importantly, we did not describe how the styles are distributed in this space. We only showed users the car's behavior for the latent values in each corner. For APReL, we explained that the interface will present two options and users must select the best option that trains the car to achieve their target style. After each selection, the interface updated its estimate of the user's style and showed the learned behavior. For both methods, users had to achieve the desired style within a tolerance of $\pm 15$ km/h speed and $\pm 5$ feet distance.

\p{Dependent Variables} 
For the indirect approach (APReL) we recorded the total number of queries that users had to answer for personalizing the car's behavior. For the direct approach (PECAN) we counted the total number of points that users had to visualize for finding their preferred style. 
\reb{
We refer to the total number of queries or clicks required to achieve the target style as \textit{Attempts}. 
Although clicks and queries represent different actions, they indicate the number of times users had to make a decision: where to click or which trajectory to choose.
We also measured the difference (i.e., Euclidean distance) between the target style and the style achieved using each approach (\textit{Style Error}). We did not compare the total time required to achieve the target style due to the fundamental differences between the two approaches. While PECAN is trained offline and instantly decodes the selected style, APReL requires significantly more time as it generates the queries and learns the styles online.

Finally, to subjectively compare the direct and indirect approaches, we collected user responses}
for the same scales as in Table~\ref{tab:questions}. We also added another scale for measuring if users perceived that they required fewer attempts (clicks or queries) to personalize the robot as they gained more experience with each method (\textit{Learn}). 

\p{Hypothesis}
We hypothesized that:
\begin{quote}
    \p{H3} \textit{Users will find it easier to personalize the driving style of the autonomous car with our direct approach (PECAN) as compared to the indirect baseline (APReL).}

    \p{H4} \textit{Users will prefer using our direct approach (PECAN) over the indirect baseline (APReL) for personalizing the driving style of the autonomous car.}
\end{quote}

\begin{figure}[t]
    \begin{center}
        \includegraphics[width=\columnwidth]{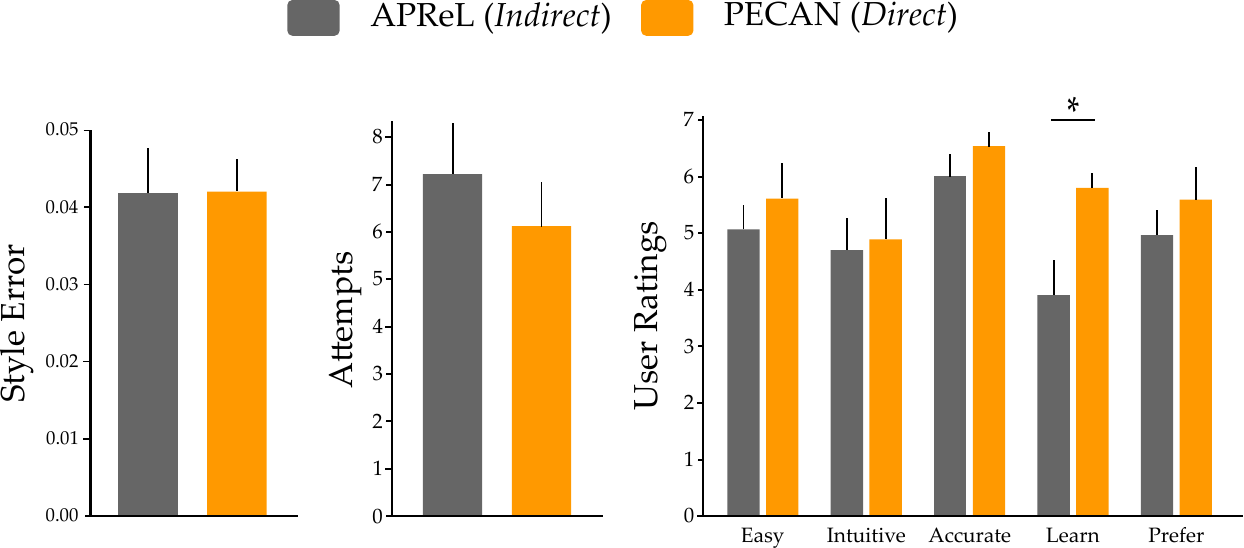}
        \caption{Objective and subjective results for the second user study. (Left) The average error in the style of the personalized car trajectory in both tasks. (Center) The total number of attempts (queries or clicks) required by users to personalize the car in both tasks. (Right) Users found both the direct and indirect approaches to be \textit{intuitive}, \textit{accurate}, and \textit{easy} to use. In particular, they perceived that they needed fewer clicks to find their desired style as they gained more experience with our direct approach. \reb{An asterisk (*) denotes statistical significance and the error bars indicate standard error.}}
        \label{fig:carlo_study}
    \end{center}
\end{figure}

\p{Results}
Our results are summarized in \fig{carlo_study}. Users were able to successfully personalize the style of the autonomous car with both direct (PECAN) and indirect (APReL) approaches. When using PECAN, $6$ out of $10$ users were able to personalize the car's style with a single click in at least one of the driving tasks! By contrast, there was only one instance when APReL learned the target style after a single query. Although users required fewer attempts (clicks or queries) on average to achieve their target style with PECAN ($M=6.1$, $SE=0.87$) than with APReL ($M=7.2$, $SE=1.05$), a two-tailed paired t-test did not show a significant difference ($t(9)=-1.14$, $p=0.28$). 

Overall, $7$ out of $10$ users stated in the survey that they preferred using our direct approach. While they gave a higher rating for PECAN ($M=5.6$, $SE=0.57$) than APReL ($M=4.9$, $SE=0.45$) on the \textit{Prefer} scale, the difference was not statistically significant. We only saw a significant difference in their subjective ratings for \textit{Learn} ($t(9)=2.67$, $p<0.05$). 

\p{Discussion}
While a majority of the users performed slightly better with our direct approach (PECAN) and preferred it over the indirect approach (APReL), the differences were not sufficient to support either hypothesis. There are two potential reasons for this result: 

First, users may have individual preferences for how they personalize the robot's behavior. For example, most users preferred PECAN because they liked that they could quickly change the car's behavior without waiting for it to learn, stating that it was ``\textit{quicker at learning and took fewer attempts}''. On the other hand, some users preferred APReL since they found it more convenient to passively respond to queries than actively selecting their style, even if it took more time, stating that they liked to ``\textit{just observe and then decide on the preferred trajectory}".

Second, the baseline had direct access to the features that define the car's style. By contrast, our approach had to learn a representation of the styles from data. This meant that while the baseline could directly reason about the car's driving style, users had to spend some time with our interface to understand how the latent values mapped to actual styles. Notably, in this study, users did not have any practice time before using the interfaces. They only interacted with the interfaces $6$-$7$ times on average. Based on the subjective responses of users for the \textit{Learn} scale, we think that people can do better with more experience using PECAN.
Hence, we conducted a follow-up study to further validate our findings and develop a better understanding of the advantages and disadvantages of using direct and indirect methods for personalizing robot behaviors.


\begin{figure}[t]
    \begin{center}
    \includegraphics[width=\columnwidth]{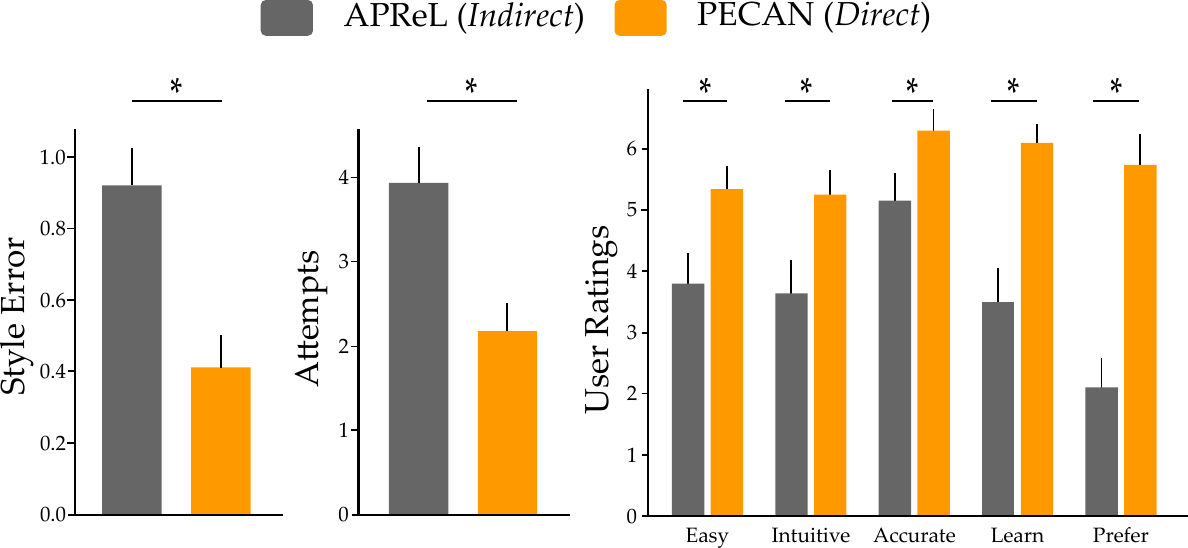}
    \caption{Objective and subjective results for the follow-up study comparing direct and indirect approaches for personalizing robot styles with experienced users. (Left) The average error in the style of the personalized car trajectory in both tasks. (Center) The average number of attempts (queries or clicks) required by users to personalize the car for each target style across both tasks. After practicing with both approaches for $10$ minutes, users were able to achieve their target style more accurately with our direct approach, PECAN, and needed fewer attempts to do so than the indirect baseline, APReL. (Right) Subjectively, experienced users found PECAN to be more \textit{intuitive}, \textit{accurate}, and \textit{easy} to use than APReL. Consequently, they preferred it over the indirect baseline for personalizing the car's trajectory across tasks. \reb{An asterisk (*) denotes statistical significance and the error bars indicate standard error.}}
    \label{fig:follow_study}
    \end{center}
\end{figure}

\subsection{Follow-up Study: Direct vs. Indirect Personalization} \label{sec:follow}

Participants in the second user study reported that they needed fewer attempts to personalize the car's behavior as they gained more experience with our direct approach. Therefore, in our follow-up study we compared our direct approach, PECAN, to the indirect baseline, APReL, with users who have practiced with both approaches. 

We recruited $10$ new participants ($1$ female, average age $20 \pm 1$ years) from the Virginia Tech community. None of these participants were involved in the first two studies. We followed the same procedure as the previous study with the following changes: (i) Before starting the experiment, we gave the participants $10$ minutes with each approach to practice personalizing the driving style of the autonomous car. (ii) To obtain a more consistent measurement of their objective performance, we asked the participants to personalize the car's trajectory for four distinct target styles, as compared to just one target style in Section~\ref{sec:indirect}. For each target, users were given a maximum of ten attempts (i.e., clicks or queries).

\p{Hypothesis} We extended \textbf{H3} and \textbf{H4} to obtain the following hypotheses:
\begin{quote}
    \p{H5} \textit{After gaining experience with both approaches, users will find it easier to personalize the autonomous car's driving style with our direct approach (PECAN) than the indirect baseline (APReL).}

    \p{H6} \textit{After gaining experience with both approaches, users will prefer using our direct approach (PECAN) to the indirect baseline (APReL) for personalizing the car's style.}
\end{quote}

\reb{
\p{Practice} Many participants demonstrated a similar approach to finding their preferred style using PECAN. They first visualized the car's trajectory for two to four points in different regions of the canonical space to understand how the latent dimensions aligned with the actual styles. Based on this initial exploration, participants chose a point that would most probably result in their desired style. If this point was still outside the allowed limits, they incrementally moved the point in a direction that would bring them closer to their target. We did not observe such guided behavior when users practiced with APReL. To assist users in recalling their initial interactions, we displayed their previous clicks in PECAN's canonical space and their previous query responses on APReL's interface. However, some users still preferred to note down the actual styles corresponding to their clicks or query responses on paper, suggesting that both interfaces could benefit from more intuitive visual cues.
}

\p{Results} The results of the follow-up study are shown in \fig{follow_study}. After practicing with both approaches, we observed that users required significantly fewer clicks or queries to personalize the autonomous car's style with PECAN than with APReL. \reb{When using PECAN, many participants were able to skip the initial exploration phase that they exhibited during practice.} A two-tailed paired t-test showed a statistically significant difference ($t(9)=-3.28$, $p < 0.01$) between the average number of attempts required by users per style with PECAN ($M=2.1$, $SE=0.33$) and APReL ($M=3.9$, $SE=0.48$). Subjectively, experienced users found our direct approach, PECAN, to be significantly \textit{easier} ($p < 0.01$) and \textit{intuitive} ($p < 0.01$) than the indirect approach, APReL. This result supported our hypothesis \textbf{H5}.

\begin{table}[t]
\caption{Open-ended responses from participants in the follow-up study that highlight the pros and cons of direct and indirect approaches for personalizing robot behaviors.}
\label{tab:follow}
\resizebox{\textwidth}{!}{%
\begin{tabular}{l|l} \hline
\multicolumn{1}{c|}{\textbf{Direct (PECAN)}} & \multicolumn{1}{c}{\textbf{Indirect (APReL)}}\\ \hline

\rule{0pt}{2.5ex}
\begin{tabular}[t]{@{}l@{}}\textbf{\textit{Pros:}}\vspace{1ex}\\
"This interface felt more \textbf{\textit{intuitive}} and made the experience\\ feel more personable."\vspace{1ex}\\
"I liked the interface as I was able to get closer to the\\ optimal point \textbf{\textit{easily and quickly}}."\vspace{1ex}\\
"I liked that it was relatively easy to place points on the\\ graph and see what reaction they had on the scenario."\vspace{1ex}\\
"Having an array of options rather than a binary decision\\ made it feel \textbf{\textit{much more personal}} and easy to use."\vspace{1ex}\end{tabular} 
& \begin{tabular}[t]{@{}l@{}} \textbf{\textit{Pros:}}\vspace{1ex}\\
"I liked how the two \textbf{\textit{options}} for trajectory that were\\ given were \textbf{\textit{different}}."\vspace{1ex}\\
"I thought that it was relatively easy to make changes."\end{tabular}\\ \hline

\rule{0pt}{2.5ex}
\begin{tabular}[t]{@{}l@{}}\textbf{\textit{Cons:}}\vspace{1ex}\\ 
"Definitely a bit tricky at first but once there were a few data\\ points to base my entries off of it became much easier."\vspace{1ex}\\
"One thing I did not like about the interface was how the \\values were not on the grid."\vspace{1ex}\\
"I did not like that the axes of the graph were \textbf{not labeled}."\vspace{1ex}\\
"The one thing I struggled with is how the grid is \textbf{\textit{not linear}}."\vspace{1ex}\end{tabular}
& \begin{tabular}[t]{@{}l@{}}\textbf{\textit{Cons:}}\vspace{1ex}\\
"It took many \textbf{\textit{more tries}} for me to get the speed\\ and distance close to the target."\vspace{1ex}\\ 
"I overall did not like this as it seemed to give me\\ substantially \textbf{\textit{less control}} over what was happening."\vspace{1ex}\\
"Sometimes I \textbf{\textit{couldnt understand}} why the\\ trajectory didnt change in the way I wanted."\vspace{1ex}\\
"I did not like that I was unable to tune speed or\\ distance independently."\vspace{1ex}\end{tabular}\\
\hline
\end{tabular}%
}
\end{table}

Overall, $9$ out of $10$ users stated that they preferred to directly specify the style instead of providing indirect feedback, and gave a significantly higher rating ($t(9)=3.98$, $p<0.01$) for PECAN ($M=5.7$, $SE=0.48$) than APReL ($M=2.1$, $SE=0.48$) on the \textit{Prefer} scale. This result supported our hypothesis \textbf{H6}.

\p{Takeaways}
In this follow-up study users had the opportunity to practice personalizing the car's driving style with both direct (PECAN) and indirect (APReL) approaches. We did not describe how the styles were distributed in the canonical space for PECAN nor did we explain how APReL learned from the user's choices. After just $10$ minutes of practice, users found PECAN to be easier and more intuitive than APReL, enabling them to achieve the target styles in fewer attempts.

Each user was tasked with personalizing the car's motion for four different target styles, allowing them to interact with each approach more times than in the previous study. Here we observed that in a few instances, APReL presented users with two trajectory options, neither of which aligned with their target style. For example, one user stated that ``\textit{often, my options were to increase my speed and distance or decrease the speed and distance when I needed to tune them opposite of each other}". Such instances were confusing for users, causing them to not achieve their target styles within ten attempts. Therefore, users had a significantly higher style error with APReL than with PECAN.

We summarize the feedback provided by users on the advantages and disadvantages of both approaches in Table~\ref{tab:follow}. We have included all comments: we only omit redundant comments and surplus details for clarity. Overall, users found that directly personalizing the car's style using PECAN was more intuitive, easy, quick, and personable. Although understanding how the latent values mapped to actual styles was initially tricky, users reported that it became much easier after a few tries. Conversely, while users appreciated the options presented by the indirect approach (APReL), they felt they had less control over its learning process and sometimes struggled to understand how their choices affected the car's learned behavior.

A common critique of our direct approach was that users wanted the styles and axes to be labeled in the canonical space. However, unlike the indirect approach, PECAN does not know the actual styles. We only use weak supervision to learn the canonical space from demonstration data. To enumerate the actual styles in the canonical space, we would need labels that specify the exact styles of trajectories in the dataset, not just whether they have similar styles. Another critique was that, although our canonical space was monotonic, users would have preferred it to be linear. This meant that, after visualizing the car's style for a couple of points in the canonical space, users could estimate the direction in which their target style would lie in the space, but not the exact distance. We aim to address this issue in future work by inducing proportionality in our canonical space. Despite this limitation, our results show that a monotonic canonical space is sufficient for users to find their desired style in a few clicks.

\reb{
\p{Applications}
Both APReL and PECAN offer distinct advantages. PECAN enables users to instantly choose their desired style, but requires them to determine the specific point that aligns with their preference. While it places some cognitive burden on users, our follow-up study indicates that this burden decreases with practice. In contrast, APReL shifts the cognitive burden to the robot, which selects the trajectories to present to the user and infers the user's preference based on their choices. This reduces the user's immediate effort but requires significantly more time and interactions for the robot to learn preferences online. Another distinction is that PECAN learns a fixed canonical space, while recent approaches have extended APReL to incorporate new features through novel forms of human demonstrations~\cite{bobu2022inducing}. In our current approach, users would need to demonstrate and label the extremes of any new style variable they wish to add to PECAN's canonical space.

Overall, we posit that PECAN is best suited for applications where users want to \textit{quickly and repeatedly change the robot's style} within a fixed spectrum. For example, users may want to incrementally make their autonomous car more aggressive or defensive depending on how late they are running. Conversely, since APReL cannot facilitate quick adjustments, it may be preferable in scenarios where users want to passively respond to the robot's queries, aiming to define the robot's style once and then repeatedly use the same style.
}
\reb{
\section{Practical Considerations}

In this section, we discuss the key design considerations for implementing and using PECAN in practice. Here we also mention PECAN's limitations and potential directions for future work.

\p{Dimensionality of canonical space}
In our experiments, we assume that the dimensionality of the styles is known and design the canonical space to have the same number of dimensions as the true styles. But more generally, the dimensionality of the canonical space does not need to match the true style variables. Rather, it depends on the number of style variables that the user wants to personalize. For example, the aggressiveness of an autonomous car can depend on multiple variables such as the car's speed, the distance it maintains from other vehicles, the number of times it changes lanes to overtake other cars, etc. However, users may not want to tune each variable individually --- they may simply care about changing the overall driving style to be more aggressive or defensive. In this case, we would only need a 1D canonical space with the most aggressive driving style at one end and the least aggressive style at the other.

Therefore, practitioners can decide the dimensionality of their canonical space based on the number of extreme styles labeled by the user. Ideally, users must label $2^{d}$ extremes for a $d$-dimensional latent space, e.g., a 3D style space would have $8$ extremes (i.e., corners of a cube). 
Using a canonical space that is smaller or larger than the exact dimensions needed to model the extreme labels can significantly affect PECAN's performance. Our experiments in Appendix~\ref{app:dimension} demonstrate that adding an extra dimension causes the latent styles to also encode task information, decreasing the consistency of the canonical space. Conversely, removing a dimension of the canonical space preserves consistency, but it prevents the styles from being monotonically arranged in the canonical space.
Future work can explore data-driven approaches that do not rely on having access to extreme styles for determining the dimensionality of the canonical space~\cite{kobayashi2023sparse}.
}

\reb{
\p{Demonstration of extreme styles} 
A key feature of PECAN is the monotonicity of its canonical space which makes the interface intuitive to use. By minimizing the combined loss in Equation~\ref{eq:combined}, PECAN encodes the extreme styles to the corners of the canonical space, while placing the intermediate styles monotonically between the extremes. To achieve this, PECAN must have access to the demonstrations and labels of all the extremes of the style spectrum. 
When demonstrations and labels for certain extremes are missing, the cross-entropy loss will still form a simplex with the available labels. For example, if the data only contains labels for two of the four extremes of a 2D style space, PECAN will encode them to opposite corners of the canonical space leaving the other two corners to potentially encode intermediate styles. Therefore, the learned canonical space may no longer be monotonic.

Our experiment in Appendix~\ref{app:dimension} demonstrates that the monotonicity of the canonical space reduces when the extreme labels do not align with the dimensionality of the canonical space. To address such cases with our current implementation, practitioners would either need to reduce the dimensionality of the canonical space to match the existing labels or provide the missing labels for extremes of the available demonstrations. 
We emphasize that labeling the style extremes is more intuitive than previous approaches which require users to quantify the robot's style~\cite{schrum2024maveric}.
Developing unsupervised methods that can automatically identify the style extremes by comparing the trajectories is an exciting direction for future work.

\p{Reconstructing intermediate styles} 
In addition to demonstrations of extreme styles, PECAN also requires unlabeled demonstrations of intermediate styles for training the decoder to accurately reconstruct trajectories from different latent values. Our preliminary tests showed that when the data does not contain a sufficient number of intermediate demonstrations, the decoder may fail to generate meaningful trajectories for certain points in the canonical space. In our experiments, we mitigated this concern by providing the minimum number of demonstrations required to train the decoder effectively. 
We found that PECAN requires more training demonstrations as the complexity of tasks and styles increases. For instance, $8$ demonstrations were sufficient to learn the manipulation tasks in our user study because the environment was static and the styles were only one dimensional. On the other hand, we needed $16$ demonstrations to learn the dynamic driving tasks which had two-dimensional styles.
We have conducted further experiments in Appendix~\ref{app:intermediate} demonstrating that PECAN's reconstructions improve as we increase the number of intermediate styles in the training data.
Future work should investigate how the decoder's output can be constrained to guarantee that the generated trajectories are safe even when trained with insufficient demonstrations.
}

\reb{
\p{Personalizing high-dimensional styles}
In this work, we evaluate PECAN across styles ranging from one to three dimensions. As explained earlier, the dimensionality of the latent styles depends on the number of variables that users choose to personalize, rather than the total number of variables that define the robot's behavior. Typically, we expect that users would only prefer personalizing a few style variables directly. However, can PECAN retain its user-friendly properties when learning canonical spaces with more than three style dimensions? To explore this, we present additional experiments in Appendix~\ref{app:generalize} testing PECAN on styles with four to six dimensions. Our results indicate that PECAN maintains its monotonicity as we increase the dimensionality of the canonical space with only a small decrease in the consistency of its canonical spaces.

Another consideration when learning high-dimensional styles is how the canonical spaces can be visualized for direct personalization. In our experiments, we present two interfaces for visualizing the canonical space: a linear slider for 1D styles and an interactive square for 2D styles, as shown in \fig{interfaces}. Practitioners can use a combination of these interfaces to represent canonical spaces with three or more dimensions. For example, we can use $n$ sliders to personalize $n$-dimensional robot styles. 
In such cases, we may want each latent dimension (i.e., slider) to correspond to a distinct style variable so that the interface is intuitive to use.

Prior research in representation learning has explored the problem of disentangling the latent variables such that each dimension represents an independent factor of variation~\cite{wang2024disentangled}. Many of these approaches use Variational Autoencoders (VAEs) to disentangle the latent dimensions in an unsupervised manner by forcing the posterior latent representation to be a multivariate Gaussian with an Identity covariance matrix~\cite{burgess2018understanding, mathieu2019disentangling}. However, prior studies have shown that while this disentangles the dimensions of the posterior samples, the dimensions of its mean (which is what users will specify when choosing their desired style) can remain correlated~\cite{locatello2019challenging}. Hence, some form of supervision is necessary to disentangle the latent dimensions into independent variables that the user wants to personalize~\cite{hristov2020disentangled}.



While our current approach obtains labels for the extremes of the style space, it does not know which extremes correspond to opposite ends of the same style dimension. As a result, the dimensions of our canonical space can get entangled. Therefore, when learning high-dimensional canonical spaces, practitioners can obtain labels for extremes of each style dimension individually. Although this requires humans to have additional domain knowledge, they would only need to label $2\times d$ extremes for a $d$-dimensional style, which is significantly fewer than the $2^{d}$ labels required in our current implementation. For example, for a 3D style $\theta$ = [$\theta_{1}, \theta_{2}, \theta_{3}$] users will need to label $6$ extreme styles, i.e., the low and high values for each dimension --- [low $\theta_1$, high $\theta_1$], [low $\theta_2$, high $\theta_2$], and [low $\theta_3$, high $\theta_3$]. In contrast, as discussed earlier, our current approach would require $8$ labels (i.e., corners of a cubic latent space).
With these labels, we can train separate 1D canonical spaces for each style variable using the same loss functions proposed in Section~\ref{sec:method}. We look forward to investigating such approaches for disentangling the latent dimensions in future work and examining whether users would find it intuitive to personalize more than three style dimensions.

\p{Other variations in trajectories}
Lastly, PECAN assumes that all tasks in the dataset share the same styles. In our simulation experiments, we found that as the tasks become more dissimilar, PECAN finds it easier to distinguish their trajectories and encode the latent task vectors. However, in real-world scenarios, when tasks become too dissimilar they may no longer have the same styles. In such cases, we may need to identify subsets of tasks that share similar styles and then learn separate canonical spaces for these subsets.

Another limitation of our approach is that we only consider variation in trajectories due to styles specified by the user. However, in many tasks, trajectories can vary due to environmental factors such as changes in object positions or the behavior of other agents. To account for such variations, we would need to condition our proposed architecture on the environment state. For example, we can provide the object positions as an additional input to the style encoder and trajectory decoder to separate the variation in styles from the environment variations in a robot manipulation task.
In this work, we assume that the environment remains consistent to emphasize the personalization of robot styles and leave extending our approach to include environment variations for future work.
}

\section{Conclusion}

In this paper we enabled humans to directly personalize robot behaviors through a canonical style space.
We first introduced PECAN, a learning and interfaces algorithm that leverages weak supervision to construct the canonical space from task demonstrations.
Next, we theoretically demonstrated why the model structure, training data, and loss functions used in PECAN help ensure that this canonical space is intuitive and user-friendly.
In practice, our approach outputs a low-dimensional manifold; each point in the manifold corresponds to a style, and humans can specify their desired style across each task in the dataset by simply clicking on their preferred point.
When experimentally compared to the alternatives, PECAN resulted in a more consistent interface that participants found easier to use over repeated interactions.

\bibliographystyle{ACM-Reference-Format}
\bibliography{references}


\begin{thebibliography}{57}


\ifx \showCODEN    \undefined \def \showCODEN     #1{\unskip}     \fi
\ifx \showDOI      \undefined \def \showDOI       #1{#1}\fi
\ifx \showISBNx    \undefined \def \showISBNx     #1{\unskip}     \fi
\ifx \showISBNxiii \undefined \def \showISBNxiii  #1{\unskip}     \fi
\ifx \showISSN     \undefined \def \showISSN      #1{\unskip}     \fi
\ifx \showLCCN     \undefined \def \showLCCN      #1{\unskip}     \fi
\ifx \shownote     \undefined \def \shownote      #1{#1}          \fi
\ifx \showarticletitle \undefined \def \showarticletitle #1{#1}   \fi
\ifx \showURL      \undefined \def \showURL       {\relax}        \fi
\providecommand\bibfield[2]{#2}
\providecommand\bibinfo[2]{#2}
\providecommand\natexlab[1]{#1}
\providecommand\showeprint[2][]{arXiv:#2}

\bibitem[\protect\citeauthoryear{Abdal, Qin, and Wonka}{Abdal et~al\mbox{.}}{2019}]%
        {abdal2019image2stylegan}
\bibfield{author}{\bibinfo{person}{Rameen Abdal}, \bibinfo{person}{Yipeng Qin}, {and} \bibinfo{person}{Peter Wonka}.} \bibinfo{year}{2019}\natexlab{}.
\newblock \showarticletitle{Image2stylegan: {H}ow to embed images into the stylegan latent space?}. In \bibinfo{booktitle}{\emph{Proceedings of the IEEE/CVF international conference on computer vision}}. \bibinfo{pages}{4432--4441}.
\newblock


\bibitem[\protect\citeauthoryear{Allshire, Mart{\'\i}n-Mart{\'\i}n, Lin, Manuel, Savarese, and Garg}{Allshire et~al\mbox{.}}{2021}]%
        {allshire2021laser}
\bibfield{author}{\bibinfo{person}{Arthur Allshire}, \bibinfo{person}{Roberto Mart{\'\i}n-Mart{\'\i}n}, \bibinfo{person}{Charles Lin}, \bibinfo{person}{Shawn Manuel}, \bibinfo{person}{Silvio Savarese}, {and} \bibinfo{person}{Animesh Garg}.} \bibinfo{year}{2021}\natexlab{}.
\newblock \showarticletitle{Laser: {L}earning a latent action space for efficient reinforcement learning}. In \bibinfo{booktitle}{\emph{2021 IEEE International Conference on Robotics and Automation (ICRA)}}. IEEE, \bibinfo{pages}{6650--6656}.
\newblock


\bibitem[\protect\citeauthoryear{Amin, Jiang, and Singh}{Amin et~al\mbox{.}}{2017}]%
        {amin2017repeated}
\bibfield{author}{\bibinfo{person}{Kareem Amin}, \bibinfo{person}{Nan Jiang}, {and} \bibinfo{person}{Satinder Singh}.} \bibinfo{year}{2017}\natexlab{}.
\newblock \showarticletitle{Repeated inverse reinforcement learning}. In \bibinfo{booktitle}{\emph{Proceedings of the 31st International Conference on Neural Information Processing Systems}}. \bibinfo{pages}{1813--1822}.
\newblock


\bibitem[\protect\citeauthoryear{B{\i}y{\i}k, Losey, Palan, Landolfi, Shevchuk, and Sadigh}{B{\i}y{\i}k et~al\mbox{.}}{2022a}]%
        {biyik2022learning}
\bibfield{author}{\bibinfo{person}{Erdem B{\i}y{\i}k}, \bibinfo{person}{Dylan~P Losey}, \bibinfo{person}{Malayandi Palan}, \bibinfo{person}{Nicholas~C Landolfi}, \bibinfo{person}{Gleb Shevchuk}, {and} \bibinfo{person}{Dorsa Sadigh}.} \bibinfo{year}{2022}\natexlab{a}.
\newblock \showarticletitle{Learning reward functions from diverse sources of human feedback: {O}ptimally integrating demonstrations and preferences}.
\newblock \bibinfo{journal}{\emph{The International Journal of Robotics Research}} \bibinfo{volume}{41}, \bibinfo{number}{1} (\bibinfo{year}{2022}), \bibinfo{pages}{45--67}.
\newblock


\bibitem[\protect\citeauthoryear{B{\i}y{\i}k, Palan, Landolfi, Losey, and Sadigh}{B{\i}y{\i}k et~al\mbox{.}}{2019}]%
        {biyik2019asking}
\bibfield{author}{\bibinfo{person}{Erdem B{\i}y{\i}k}, \bibinfo{person}{Malayandi Palan}, \bibinfo{person}{Nicholas~C Landolfi}, \bibinfo{person}{Dylan~P Losey}, {and} \bibinfo{person}{Dorsa Sadigh}.} \bibinfo{year}{2019}\natexlab{}.
\newblock \showarticletitle{Asking easy questions: {A} user-friendly approach to active reward learning}. In \bibinfo{booktitle}{\emph{Annual Conference on Robot Learning}}. \bibinfo{pages}{1177--1190}.
\newblock


\bibitem[\protect\citeauthoryear{B{\i}y{\i}k, Talati, and Sadigh}{B{\i}y{\i}k et~al\mbox{.}}{2022b}]%
        {biyik2022APReL}
\bibfield{author}{\bibinfo{person}{Erdem B{\i}y{\i}k}, \bibinfo{person}{Aditi Talati}, {and} \bibinfo{person}{Dorsa Sadigh}.} \bibinfo{year}{2022}\natexlab{b}.
\newblock \showarticletitle{Aprel: {A} library for active preference-based reward learning algorithms}. In \bibinfo{booktitle}{\emph{2022 17th ACM/IEEE International Conference on Human-Robot Interaction (HRI)}}. IEEE, \bibinfo{pages}{613--617}.
\newblock


\bibitem[\protect\citeauthoryear{Bobu, Peng, Agrawal, Shah, and Dragan}{Bobu et~al\mbox{.}}{2024}]%
        {bobu2024aligning}
\bibfield{author}{\bibinfo{person}{Andreea Bobu}, \bibinfo{person}{Andi Peng}, \bibinfo{person}{Pulkit Agrawal}, \bibinfo{person}{Julie~A Shah}, {and} \bibinfo{person}{Anca~D Dragan}.} \bibinfo{year}{2024}\natexlab{}.
\newblock \showarticletitle{Aligning human and robot representations}. In \bibinfo{booktitle}{\emph{Proceedings of the 2024 ACM/IEEE International Conference on Human-Robot Interaction}}. \bibinfo{pages}{42--54}.
\newblock


\bibitem[\protect\citeauthoryear{Bobu, Wiggert, Tomlin, and Dragan}{Bobu et~al\mbox{.}}{2022}]%
        {bobu2022inducing}
\bibfield{author}{\bibinfo{person}{Andreea Bobu}, \bibinfo{person}{Marius Wiggert}, \bibinfo{person}{Claire Tomlin}, {and} \bibinfo{person}{Anca~D Dragan}.} \bibinfo{year}{2022}\natexlab{}.
\newblock \showarticletitle{Inducing structure in reward learning by learning features}.
\newblock \bibinfo{journal}{\emph{The International Journal of Robotics Research}} \bibinfo{volume}{41}, \bibinfo{number}{5} (\bibinfo{year}{2022}), \bibinfo{pages}{497--518}.
\newblock


\bibitem[\protect\citeauthoryear{Bonyani, Soleymani, and Wang}{Bonyani et~al\mbox{.}}{2024}]%
        {bonyani2024style}
\bibfield{author}{\bibinfo{person}{Mahdi Bonyani}, \bibinfo{person}{Maryam Soleymani}, {and} \bibinfo{person}{Chao Wang}.} \bibinfo{year}{2024}\natexlab{}.
\newblock \showarticletitle{Style-Based Reinforcement Learning: {T}ask Decoupling Personalization for Human-Robot Collaboration}. In \bibinfo{booktitle}{\emph{International Conference on Human-Computer Interaction}}. Springer, \bibinfo{pages}{197--212}.
\newblock


\bibitem[\protect\citeauthoryear{Burgess, Higgins, Pal, Matthey, Watters, Desjardins, and Lerchner}{Burgess et~al\mbox{.}}{2018}]%
        {burgess2018understanding}
\bibfield{author}{\bibinfo{person}{Christopher~P Burgess}, \bibinfo{person}{Irina Higgins}, \bibinfo{person}{Arka Pal}, \bibinfo{person}{Loic Matthey}, \bibinfo{person}{Nick Watters}, \bibinfo{person}{Guillaume Desjardins}, {and} \bibinfo{person}{Alexander Lerchner}.} \bibinfo{year}{2018}\natexlab{}.
\newblock \showarticletitle{Understanding disentangling in $\beta$-VAE}.
\newblock \bibinfo{journal}{\emph{arXiv preprint arXiv:1804.03599}} (\bibinfo{year}{2018}).
\newblock


\bibitem[\protect\citeauthoryear{Cao, Biyik, Wang, Raventos, Gaidon, Rosman, and Sadigh}{Cao et~al\mbox{.}}{2020}]%
        {cao2020reinforcement}
\bibfield{author}{\bibinfo{person}{Zhangjie Cao}, \bibinfo{person}{Erdem Biyik}, \bibinfo{person}{Woodrow~Z. Wang}, \bibinfo{person}{Allan Raventos}, \bibinfo{person}{Adrien Gaidon}, \bibinfo{person}{Guy Rosman}, {and} \bibinfo{person}{Dorsa Sadigh}.} \bibinfo{year}{2020}\natexlab{}.
\newblock \showarticletitle{Reinforcement Learning based Control of Imitative Policies for Near-Accident Driving}. In \bibinfo{booktitle}{\emph{Proceedings of Robotics: {S}cience and Systems (RSS)}}.
\newblock


\bibitem[\protect\citeauthoryear{Christiano, Leike, Brown, Martic, Legg, and Amodei}{Christiano et~al\mbox{.}}{2017}]%
        {christiano2017deep}
\bibfield{author}{\bibinfo{person}{Paul~F Christiano}, \bibinfo{person}{Jan Leike}, \bibinfo{person}{Tom Brown}, \bibinfo{person}{Miljan Martic}, \bibinfo{person}{Shane Legg}, {and} \bibinfo{person}{Dario Amodei}.} \bibinfo{year}{2017}\natexlab{}.
\newblock \showarticletitle{Deep reinforcement learning from human preferences}.
\newblock \bibinfo{journal}{\emph{Advances in neural information processing systems}} \bibinfo{volume}{30}, \bibinfo{number}{9} (\bibinfo{year}{2017}), \bibinfo{pages}{4302--4310}.
\newblock


\bibitem[\protect\citeauthoryear{Co-Reyes, Liu, Gupta, Eysenbach, Abbeel, and Levine}{Co-Reyes et~al\mbox{.}}{2018}]%
        {co2018self}
\bibfield{author}{\bibinfo{person}{John Co-Reyes}, \bibinfo{person}{YuXuan Liu}, \bibinfo{person}{Abhishek Gupta}, \bibinfo{person}{Benjamin Eysenbach}, \bibinfo{person}{Pieter Abbeel}, {and} \bibinfo{person}{Sergey Levine}.} \bibinfo{year}{2018}\natexlab{}.
\newblock \showarticletitle{Self-consistent trajectory autoencoder: Hierarchical reinforcement learning with trajectory embeddings}. In \bibinfo{booktitle}{\emph{International conference on machine learning}}. PMLR, \bibinfo{pages}{1009--1018}.
\newblock


\bibitem[\protect\citeauthoryear{Cohen, Huang, Chen, Benesty, Benesty, Chen, Huang, and Cohen}{Cohen et~al\mbox{.}}{2009}]%
        {cohen2009pearson}
\bibfield{author}{\bibinfo{person}{Israel Cohen}, \bibinfo{person}{Yiteng Huang}, \bibinfo{person}{Jingdong Chen}, \bibinfo{person}{Jacob Benesty}, \bibinfo{person}{Jacob Benesty}, \bibinfo{person}{Jingdong Chen}, \bibinfo{person}{Yiteng Huang}, {and} \bibinfo{person}{Israel Cohen}.} \bibinfo{year}{2009}\natexlab{}.
\newblock \showarticletitle{Pearson correlation coefficient}.
\newblock \bibinfo{journal}{\emph{Noise reduction in speech processing}} (\bibinfo{year}{2009}), \bibinfo{pages}{1--4}.
\newblock


\bibitem[\protect\citeauthoryear{Dupont}{Dupont}{2018}]%
        {dupont2018learning}
\bibfield{author}{\bibinfo{person}{Emilien Dupont}.} \bibinfo{year}{2018}\natexlab{}.
\newblock \showarticletitle{Learning disentangled joint continuous and discrete representations}. In \bibinfo{booktitle}{\emph{Proceedings of the 32nd International Conference on Neural Information Processing Systems}}. \bibinfo{pages}{708--718}.
\newblock


\bibitem[\protect\citeauthoryear{Graf, Hofer, Niethammer, and Kwitt}{Graf et~al\mbox{.}}{2021}]%
        {graf2021dissecting}
\bibfield{author}{\bibinfo{person}{Florian Graf}, \bibinfo{person}{Christoph Hofer}, \bibinfo{person}{Marc Niethammer}, {and} \bibinfo{person}{Roland Kwitt}.} \bibinfo{year}{2021}\natexlab{}.
\newblock \showarticletitle{Dissecting supervised contrastive learning}. In \bibinfo{booktitle}{\emph{International Conference on Machine Learning}}. PMLR, \bibinfo{pages}{3821--3830}.
\newblock


\bibitem[\protect\citeauthoryear{Guo, Jena, Hughes, Lewis, and Sycara}{Guo et~al\mbox{.}}{2021}]%
        {guo2021transfer}
\bibfield{author}{\bibinfo{person}{Yue Guo}, \bibinfo{person}{Rohit Jena}, \bibinfo{person}{Dana Hughes}, \bibinfo{person}{Michael Lewis}, {and} \bibinfo{person}{Katia Sycara}.} \bibinfo{year}{2021}\natexlab{}.
\newblock \showarticletitle{Transfer learning for human navigation and triage strategies prediction in a simulated urban search and rescue task}. In \bibinfo{booktitle}{\emph{2021 30th IEEE International Conference on Robot \& Human Interactive Communication (RO-MAN)}}. IEEE, \bibinfo{pages}{784--791}.
\newblock


\bibitem[\protect\citeauthoryear{Hejna~III and Sadigh}{Hejna~III and Sadigh}{2023}]%
        {hejna2023few}
\bibfield{author}{\bibinfo{person}{Donald~Joseph Hejna~III} {and} \bibinfo{person}{Dorsa Sadigh}.} \bibinfo{year}{2023}\natexlab{}.
\newblock \showarticletitle{Few-shot preference learning for human-in-the-loop rl}. In \bibinfo{booktitle}{\emph{Conference on Robot Learning}}. PMLR, \bibinfo{pages}{2014--2025}.
\newblock


\bibitem[\protect\citeauthoryear{Hristov, Angelov, Burke, Lascarides, and Ramamoorthy}{Hristov et~al\mbox{.}}{2020}]%
        {hristov2020disentangled}
\bibfield{author}{\bibinfo{person}{Yordan Hristov}, \bibinfo{person}{Daniel Angelov}, \bibinfo{person}{Michael Burke}, \bibinfo{person}{Alex Lascarides}, {and} \bibinfo{person}{Subramanian Ramamoorthy}.} \bibinfo{year}{2020}\natexlab{}.
\newblock \showarticletitle{Disentangled relational representations for explaining and learning from demonstration}. In \bibinfo{booktitle}{\emph{Conference on Robot Learning}}. PMLR, \bibinfo{pages}{870--884}.
\newblock


\bibitem[\protect\citeauthoryear{Jain, Sharma, Joachims, and Saxena}{Jain et~al\mbox{.}}{2015}]%
        {jain2015learning}
\bibfield{author}{\bibinfo{person}{Ashesh Jain}, \bibinfo{person}{Shikhar Sharma}, \bibinfo{person}{Thorsten Joachims}, {and} \bibinfo{person}{Ashutosh Saxena}.} \bibinfo{year}{2015}\natexlab{}.
\newblock \showarticletitle{Learning preferences for manipulation tasks from online coactive feedback}.
\newblock \bibinfo{journal}{\emph{The International Journal of Robotics Research}} \bibinfo{volume}{34}, \bibinfo{number}{10} (\bibinfo{year}{2015}), \bibinfo{pages}{1296--1313}.
\newblock


\bibitem[\protect\citeauthoryear{Jang, Gu, and Poole}{Jang et~al\mbox{.}}{2017}]%
        {jang2017categorical}
\bibfield{author}{\bibinfo{person}{Eric Jang}, \bibinfo{person}{Shixiang Gu}, {and} \bibinfo{person}{Ben Poole}.} \bibinfo{year}{2017}\natexlab{}.
\newblock \showarticletitle{Categorical Reparameterization with Gumbel-Softmax}. In \bibinfo{booktitle}{\emph{International Conference on Learning Representations}}.
\newblock


\bibitem[\protect\citeauthoryear{Jiao, Liu, Zheng, Liang, and Zhu}{Jiao et~al\mbox{.}}{2022}]%
        {jiao2022tae}
\bibfield{author}{\bibinfo{person}{Ruochen Jiao}, \bibinfo{person}{Xiangguo Liu}, \bibinfo{person}{Bowen Zheng}, \bibinfo{person}{Dave Liang}, {and} \bibinfo{person}{Qi Zhu}.} \bibinfo{year}{2022}\natexlab{}.
\newblock \showarticletitle{Tae: A semi-supervised controllable behavior-aware trajectory generator and predictor}. In \bibinfo{booktitle}{\emph{2022 IEEE/RSJ International Conference on Intelligent Robots and Systems (IROS)}}. IEEE, \bibinfo{pages}{12534--12541}.
\newblock


\bibitem[\protect\citeauthoryear{Jing, Yang, Feng, Ye, Yu, and Song}{Jing et~al\mbox{.}}{2019}]%
        {jing2019neural}
\bibfield{author}{\bibinfo{person}{Yongcheng Jing}, \bibinfo{person}{Yezhou Yang}, \bibinfo{person}{Zunlei Feng}, \bibinfo{person}{Jingwen Ye}, \bibinfo{person}{Yizhou Yu}, {and} \bibinfo{person}{Mingli Song}.} \bibinfo{year}{2019}\natexlab{}.
\newblock \showarticletitle{Neural style transfer: {A} review}.
\newblock \bibinfo{journal}{\emph{IEEE transactions on visualization and computer graphics}} \bibinfo{volume}{26}, \bibinfo{number}{11} (\bibinfo{year}{2019}), \bibinfo{pages}{3365--3385}.
\newblock


\bibitem[\protect\citeauthoryear{Karras, Laine, and Aila}{Karras et~al\mbox{.}}{2019}]%
        {karras2019stylegen}
\bibfield{author}{\bibinfo{person}{Tero Karras}, \bibinfo{person}{Samuli Laine}, {and} \bibinfo{person}{Timo Aila}.} \bibinfo{year}{2019}\natexlab{}.
\newblock \showarticletitle{A style-based generator architecture for generative adversarial networks}. In \bibinfo{booktitle}{\emph{Proceedings of the IEEE/CVF Conference on Computer Vision and Pattern Recognition (CVPR)}}.
\newblock


\bibitem[\protect\citeauthoryear{Katz, Maleki, B{\i}y{\i}k, and Kochenderfer}{Katz et~al\mbox{.}}{2021}]%
        {katz2021preference}
\bibfield{author}{\bibinfo{person}{Sydney~M Katz}, \bibinfo{person}{Amir Maleki}, \bibinfo{person}{Erdem B{\i}y{\i}k}, {and} \bibinfo{person}{Mykel~J Kochenderfer}.} \bibinfo{year}{2021}\natexlab{}.
\newblock \showarticletitle{Preference-based learning of reward function features}.
\newblock \bibinfo{journal}{\emph{arXiv preprint arXiv:2103.02727}} (\bibinfo{year}{2021}).
\newblock


\bibitem[\protect\citeauthoryear{Kobayashi and Watanuki}{Kobayashi and Watanuki}{2023}]%
        {kobayashi2023sparse}
\bibfield{author}{\bibinfo{person}{Taisuke Kobayashi} {and} \bibinfo{person}{Ryoma Watanuki}.} \bibinfo{year}{2023}\natexlab{}.
\newblock \showarticletitle{Sparse representation learning with modified q-VAE towards minimal realization of world model}.
\newblock \bibinfo{journal}{\emph{Advanced Robotics}} \bibinfo{volume}{37}, \bibinfo{number}{13} (\bibinfo{year}{2023}), \bibinfo{pages}{807--827}.
\newblock


\bibitem[\protect\citeauthoryear{Koppol, Admoni, and Simmons}{Koppol et~al\mbox{.}}{2021}]%
        {koppol2021interaction}
\bibfield{author}{\bibinfo{person}{Pallavi Koppol}, \bibinfo{person}{Henny Admoni}, {and} \bibinfo{person}{Reid~G Simmons}.} \bibinfo{year}{2021}\natexlab{}.
\newblock \showarticletitle{Interaction considerations in learning from humans}. In \bibinfo{booktitle}{\emph{IJCAI}}. \bibinfo{pages}{283--291}.
\newblock


\bibitem[\protect\citeauthoryear{Li, Canberk, Losey, and Sadigh}{Li et~al\mbox{.}}{2021}]%
        {li2021learning}
\bibfield{author}{\bibinfo{person}{Mengxi Li}, \bibinfo{person}{Alper Canberk}, \bibinfo{person}{Dylan~P Losey}, {and} \bibinfo{person}{Dorsa Sadigh}.} \bibinfo{year}{2021}\natexlab{}.
\newblock \showarticletitle{Learning human objectives from sequences of physical corrections}. In \bibinfo{booktitle}{\emph{2021 IEEE International Conference on Robotics and Automation (ICRA)}}. IEEE, \bibinfo{pages}{2877--2883}.
\newblock


\bibitem[\protect\citeauthoryear{Locatello, Bauer, Lucic, Raetsch, Gelly, Sch{\"o}lkopf, and Bachem}{Locatello et~al\mbox{.}}{2019}]%
        {locatello2019challenging}
\bibfield{author}{\bibinfo{person}{Francesco Locatello}, \bibinfo{person}{Stefan Bauer}, \bibinfo{person}{Mario Lucic}, \bibinfo{person}{Gunnar Raetsch}, \bibinfo{person}{Sylvain Gelly}, \bibinfo{person}{Bernhard Sch{\"o}lkopf}, {and} \bibinfo{person}{Olivier Bachem}.} \bibinfo{year}{2019}\natexlab{}.
\newblock \showarticletitle{Challenging common assumptions in the unsupervised learning of disentangled representations}. In \bibinfo{booktitle}{\emph{international conference on machine learning}}. PMLR, \bibinfo{pages}{4114--4124}.
\newblock


\bibitem[\protect\citeauthoryear{Losey, Bajcsy, O’Malley, and Dragan}{Losey et~al\mbox{.}}{2022}]%
        {losey2022physical}
\bibfield{author}{\bibinfo{person}{Dylan~P Losey}, \bibinfo{person}{Andrea Bajcsy}, \bibinfo{person}{Marcia~K O’Malley}, {and} \bibinfo{person}{Anca~D Dragan}.} \bibinfo{year}{2022}\natexlab{}.
\newblock \showarticletitle{Physical interaction as communication: {L}earning robot objectives online from human corrections}.
\newblock \bibinfo{journal}{\emph{The International Journal of Robotics Research}} \bibinfo{volume}{41}, \bibinfo{number}{1} (\bibinfo{year}{2022}), \bibinfo{pages}{20--44}.
\newblock


\bibitem[\protect\citeauthoryear{Lynch, Khansari, Xiao, Kumar, Tompson, Levine, and Sermanet}{Lynch et~al\mbox{.}}{2020}]%
        {lynch2020learning}
\bibfield{author}{\bibinfo{person}{Corey Lynch}, \bibinfo{person}{Mohi Khansari}, \bibinfo{person}{Ted Xiao}, \bibinfo{person}{Vikash Kumar}, \bibinfo{person}{Jonathan Tompson}, \bibinfo{person}{Sergey Levine}, {and} \bibinfo{person}{Pierre Sermanet}.} \bibinfo{year}{2020}\natexlab{}.
\newblock \showarticletitle{Learning latent plans from play}. In \bibinfo{booktitle}{\emph{Conference on robot learning}}. PMLR, \bibinfo{pages}{1113--1132}.
\newblock


\bibitem[\protect\citeauthoryear{Mandi, Liu, Lee, and Abbeel}{Mandi et~al\mbox{.}}{2022}]%
        {mandi2022towards}
\bibfield{author}{\bibinfo{person}{Zhao Mandi}, \bibinfo{person}{Fangchen Liu}, \bibinfo{person}{Kimin Lee}, {and} \bibinfo{person}{Pieter Abbeel}.} \bibinfo{year}{2022}\natexlab{}.
\newblock \showarticletitle{Towards more generalizable one-shot visual imitation learning}. In \bibinfo{booktitle}{\emph{2022 International Conference on Robotics and Automation (ICRA)}}. IEEE, \bibinfo{pages}{2434--2444}.
\newblock


\bibitem[\protect\citeauthoryear{Mathieu, Rainforth, Siddharth, and Teh}{Mathieu et~al\mbox{.}}{2019}]%
        {mathieu2019disentangling}
\bibfield{author}{\bibinfo{person}{Emile Mathieu}, \bibinfo{person}{Tom Rainforth}, \bibinfo{person}{Nana Siddharth}, {and} \bibinfo{person}{Yee~Whye Teh}.} \bibinfo{year}{2019}\natexlab{}.
\newblock \showarticletitle{Disentangling disentanglement in variational autoencoders}. In \bibinfo{booktitle}{\emph{International conference on machine learning}}. PMLR, \bibinfo{pages}{4402--4412}.
\newblock


\bibitem[\protect\citeauthoryear{Mehta and Losey}{Mehta and Losey}{2023}]%
        {mehta2023unified}
\bibfield{author}{\bibinfo{person}{Shaunak~A Mehta} {and} \bibinfo{person}{Dylan~P Losey}.} \bibinfo{year}{2023}\natexlab{}.
\newblock \showarticletitle{Unified learning from demonstrations, corrections, and preferences during physical human-robot interaction}.
\newblock \bibinfo{journal}{\emph{ACM Transactions on Human-Robot Interaction}} (\bibinfo{year}{2023}).
\newblock


\bibitem[\protect\citeauthoryear{Munzer, Toussaint, and Lopes}{Munzer et~al\mbox{.}}{2017}]%
        {munzer2017preference}
\bibfield{author}{\bibinfo{person}{Thibaut Munzer}, \bibinfo{person}{Marc Toussaint}, {and} \bibinfo{person}{Manuel Lopes}.} \bibinfo{year}{2017}\natexlab{}.
\newblock \showarticletitle{Preference learning on the execution of collaborative human-robot tasks}. In \bibinfo{booktitle}{\emph{IEEE International Conference on Robotics and Automation}}. \bibinfo{pages}{879--885}.
\newblock


\bibitem[\protect\citeauthoryear{Nikolaidis, Ramakrishnan, Gu, and Shah}{Nikolaidis et~al\mbox{.}}{2015}]%
        {nikolaidis2015efficient}
\bibfield{author}{\bibinfo{person}{Stefanos Nikolaidis}, \bibinfo{person}{Ramya Ramakrishnan}, \bibinfo{person}{Keren Gu}, {and} \bibinfo{person}{Julie Shah}.} \bibinfo{year}{2015}\natexlab{}.
\newblock \showarticletitle{Efficient model learning from joint-action demonstrations for human-robot collaborative tasks}. In \bibinfo{booktitle}{\emph{ACM/IEEE International Conference on Human-Robot Interaction}}. \bibinfo{pages}{189--196}.
\newblock


\bibitem[\protect\citeauthoryear{Osa and Ikemoto}{Osa and Ikemoto}{2020}]%
        {osa2020goal}
\bibfield{author}{\bibinfo{person}{Takayuki Osa} {and} \bibinfo{person}{Shuehi Ikemoto}.} \bibinfo{year}{2020}\natexlab{}.
\newblock \showarticletitle{Goal-conditioned variational autoencoder trajectory primitives with continuous and discrete latent codes}.
\newblock \bibinfo{journal}{\emph{SN Computer Science}} \bibinfo{volume}{1}, \bibinfo{number}{5} (\bibinfo{year}{2020}), \bibinfo{pages}{303}.
\newblock


\bibitem[\protect\citeauthoryear{Rahmatizadeh, Abolghasemi, B{\"o}l{\"o}ni, and Levine}{Rahmatizadeh et~al\mbox{.}}{2018}]%
        {rahmatizadeh2018vision}
\bibfield{author}{\bibinfo{person}{Rouhollah Rahmatizadeh}, \bibinfo{person}{Pooya Abolghasemi}, \bibinfo{person}{Ladislau B{\"o}l{\"o}ni}, {and} \bibinfo{person}{Sergey Levine}.} \bibinfo{year}{2018}\natexlab{}.
\newblock \showarticletitle{Vision-based multi-task manipulation for inexpensive robots using end-to-end learning from demonstration}. In \bibinfo{booktitle}{\emph{2018 IEEE international conference on robotics and automation (ICRA)}}. IEEE, \bibinfo{pages}{3758--3765}.
\newblock


\bibitem[\protect\citeauthoryear{Rangwani, Bansal, Sharma, Karmali, Jampani, and Babu}{Rangwani et~al\mbox{.}}{2023}]%
        {rangwani2023noisytwins}
\bibfield{author}{\bibinfo{person}{Harsh Rangwani}, \bibinfo{person}{Lavish Bansal}, \bibinfo{person}{Kartik Sharma}, \bibinfo{person}{Tejan Karmali}, \bibinfo{person}{Varun Jampani}, {and} \bibinfo{person}{R~Venkatesh Babu}.} \bibinfo{year}{2023}\natexlab{}.
\newblock \showarticletitle{Noisytwins: {C}lass-consistent and diverse image generation through stylegans}. In \bibinfo{booktitle}{\emph{Proceedings of the IEEE/CVF Conference on Computer Vision and Pattern Recognition}}. \bibinfo{pages}{5987--5996}.
\newblock


\bibitem[\protect\citeauthoryear{Ravichandar, Polydoros, Chernova, and Billard}{Ravichandar et~al\mbox{.}}{2020}]%
        {ravichandar2020recent}
\bibfield{author}{\bibinfo{person}{Harish Ravichandar}, \bibinfo{person}{Athanasios~S Polydoros}, \bibinfo{person}{Sonia Chernova}, {and} \bibinfo{person}{Aude Billard}.} \bibinfo{year}{2020}\natexlab{}.
\newblock \showarticletitle{Recent advances in robot learning from demonstration}.
\newblock \bibinfo{journal}{\emph{Annual review of control, robotics, and autonomous systems}}  \bibinfo{volume}{3} (\bibinfo{year}{2020}), \bibinfo{pages}{297--330}.
\newblock


\bibitem[\protect\citeauthoryear{Rosbach, James, Gro{\ss}johann, Homoceanu, and Roth}{Rosbach et~al\mbox{.}}{2019}]%
        {rosbach2019driving}
\bibfield{author}{\bibinfo{person}{Sascha Rosbach}, \bibinfo{person}{Vinit James}, \bibinfo{person}{Simon Gro{\ss}johann}, \bibinfo{person}{Silviu Homoceanu}, {and} \bibinfo{person}{Stefan Roth}.} \bibinfo{year}{2019}\natexlab{}.
\newblock \showarticletitle{Driving with style: {I}nverse reinforcement learning in general-purpose planning for automated driving}. In \bibinfo{booktitle}{\emph{IEEE/RSJ International Conference on Intelligent Robots and Systems}}. \bibinfo{pages}{2658--2665}.
\newblock


\bibitem[\protect\citeauthoryear{Sadigh, Dragan, Sastry, and Seshia}{Sadigh et~al\mbox{.}}{2017}]%
        {sadigh2017active}
\bibfield{author}{\bibinfo{person}{Dorsa Sadigh}, \bibinfo{person}{Anca Dragan}, \bibinfo{person}{Shankar Sastry}, {and} \bibinfo{person}{Sanjit Seshia}.} \bibinfo{year}{2017}\natexlab{}.
\newblock \showarticletitle{Active preference-based learning of reward functions}. In \bibinfo{booktitle}{\emph{Proceedings of Robotics: {S}cience and Systems (RSS)}}.
\newblock


\bibitem[\protect\citeauthoryear{Schrum, Sumner, Gombolay, and Best}{Schrum et~al\mbox{.}}{2024}]%
        {schrum2024maveric}
\bibfield{author}{\bibinfo{person}{Mariah~L Schrum}, \bibinfo{person}{Emily Sumner}, \bibinfo{person}{Matthew~C Gombolay}, {and} \bibinfo{person}{Andrew Best}.} \bibinfo{year}{2024}\natexlab{}.
\newblock \showarticletitle{Maveric: A data-driven approach to personalized autonomous driving}.
\newblock \bibinfo{journal}{\emph{IEEE Transactions on Robotics}} (\bibinfo{year}{2024}).
\newblock


\bibitem[\protect\citeauthoryear{Singh, Jang, Irpan, Kappler, Dalal, Levinev, Khansari, and Finn}{Singh et~al\mbox{.}}{2020}]%
        {singh2020scalable}
\bibfield{author}{\bibinfo{person}{Avi Singh}, \bibinfo{person}{Eric Jang}, \bibinfo{person}{Alexander Irpan}, \bibinfo{person}{Daniel Kappler}, \bibinfo{person}{Murtaza Dalal}, \bibinfo{person}{Sergey Levinev}, \bibinfo{person}{Mohi Khansari}, {and} \bibinfo{person}{Chelsea Finn}.} \bibinfo{year}{2020}\natexlab{}.
\newblock \showarticletitle{Scalable multi-task imitation learning with autonomous improvement}. In \bibinfo{booktitle}{\emph{2020 IEEE International Conference on Robotics and Automation (ICRA)}}. IEEE, \bibinfo{pages}{2167--2173}.
\newblock


\bibitem[\protect\citeauthoryear{{\'S}mieja, Wo{\l}czyk, Tabor, and Geiger}{{\'S}mieja et~al\mbox{.}}{2020}]%
        {smieja2020segma}
\bibfield{author}{\bibinfo{person}{Marek {\'S}mieja}, \bibinfo{person}{Maciej Wo{\l}czyk}, \bibinfo{person}{Jacek Tabor}, {and} \bibinfo{person}{Bernhard~C Geiger}.} \bibinfo{year}{2020}\natexlab{}.
\newblock \showarticletitle{Segma: {S}emi-supervised Gaussian mixture autoencoder}.
\newblock \bibinfo{journal}{\emph{IEEE transactions on neural networks and learning systems}} \bibinfo{volume}{32}, \bibinfo{number}{9} (\bibinfo{year}{2020}), \bibinfo{pages}{3930--3941}.
\newblock


\bibitem[\protect\citeauthoryear{Spencer, Choudhury, Barnes, Schmittle, Chiang, Ramadge, and Srinivasa}{Spencer et~al\mbox{.}}{2022}]%
        {spencer2022expert}
\bibfield{author}{\bibinfo{person}{Jonathan Spencer}, \bibinfo{person}{Sanjiban Choudhury}, \bibinfo{person}{Matthew Barnes}, \bibinfo{person}{Matthew Schmittle}, \bibinfo{person}{Mung Chiang}, \bibinfo{person}{Peter Ramadge}, {and} \bibinfo{person}{Sidd Srinivasa}.} \bibinfo{year}{2022}\natexlab{}.
\newblock \showarticletitle{Expert intervention learning: {A}n online framework for robot learning from explicit and implicit human feedback}.
\newblock \bibinfo{journal}{\emph{Autonomous Robots}} (\bibinfo{year}{2022}), \bibinfo{pages}{1--15}.
\newblock


\bibitem[\protect\citeauthoryear{Sun, Kennedy, Zhan, Anderson, Yue, and Perona}{Sun et~al\mbox{.}}{2021}]%
        {sun2021task}
\bibfield{author}{\bibinfo{person}{Jennifer~J Sun}, \bibinfo{person}{Ann Kennedy}, \bibinfo{person}{Eric Zhan}, \bibinfo{person}{David~J Anderson}, \bibinfo{person}{Yisong Yue}, {and} \bibinfo{person}{Pietro Perona}.} \bibinfo{year}{2021}\natexlab{}.
\newblock \showarticletitle{Task programming: Learning data efficient behavior representations}. In \bibinfo{booktitle}{\emph{Proceedings of the IEEE/CVF Conference on Computer Vision and Pattern Recognition}}. \bibinfo{pages}{2876--2885}.
\newblock


\bibitem[\protect\citeauthoryear{Vowels, Camgoz, and Bowden}{Vowels et~al\mbox{.}}{2020}]%
        {vowels2020gated}
\bibfield{author}{\bibinfo{person}{Matthew~J Vowels}, \bibinfo{person}{Necati~Cihan Camgoz}, {and} \bibinfo{person}{Richard Bowden}.} \bibinfo{year}{2020}\natexlab{}.
\newblock \showarticletitle{Gated variational autoencoders: {I}ncorporating weak supervision to encourage disentanglement}. In \bibinfo{booktitle}{\emph{2020 15th IEEE International Conference on Automatic Face and Gesture Recognition (FG 2020)}}. IEEE, \bibinfo{pages}{125--132}.
\newblock


\bibitem[\protect\citeauthoryear{Wang, Chen, Wu, Zhu, et~al\mbox{.}}{Wang et~al\mbox{.}}{2024}]%
        {wang2024disentangled}
\bibfield{author}{\bibinfo{person}{Xin Wang}, \bibinfo{person}{Hong Chen}, \bibinfo{person}{Zihao Wu}, \bibinfo{person}{Wenwu Zhu}, {et~al\mbox{.}}} \bibinfo{year}{2024}\natexlab{}.
\newblock \showarticletitle{Disentangled representation learning}.
\newblock \bibinfo{journal}{\emph{IEEE Transactions on Pattern Analysis and Machine Intelligence}} (\bibinfo{year}{2024}).
\newblock


\bibitem[\protect\citeauthoryear{Wang, Lee, Hakhamaneshi, Abbeel, and Laskin}{Wang et~al\mbox{.}}{2022}]%
        {wang2022skill}
\bibfield{author}{\bibinfo{person}{Xiaofei Wang}, \bibinfo{person}{Kimin Lee}, \bibinfo{person}{Kourosh Hakhamaneshi}, \bibinfo{person}{Pieter Abbeel}, {and} \bibinfo{person}{Michael Laskin}.} \bibinfo{year}{2022}\natexlab{}.
\newblock \showarticletitle{Skill preferences: Learning to extract and execute robotic skills from human feedback}. In \bibinfo{booktitle}{\emph{Conference on Robot Learning}}. PMLR, \bibinfo{pages}{1259--1268}.
\newblock


\bibitem[\protect\citeauthoryear{Wilde, Blidaru, Smith, and Kuli{\'c}}{Wilde et~al\mbox{.}}{2020}]%
        {wilde2020improving}
\bibfield{author}{\bibinfo{person}{Nils Wilde}, \bibinfo{person}{Alexandru Blidaru}, \bibinfo{person}{Stephen~L Smith}, {and} \bibinfo{person}{Dana Kuli{\'c}}.} \bibinfo{year}{2020}\natexlab{}.
\newblock \showarticletitle{Improving user specifications for robot behavior through active preference learning: {F}ramework and evaluation}.
\newblock \bibinfo{journal}{\emph{The International Journal of Robotics Research}} \bibinfo{volume}{39}, \bibinfo{number}{6} (\bibinfo{year}{2020}), \bibinfo{pages}{651--667}.
\newblock


\bibitem[\protect\citeauthoryear{Woodworth, Ferrari, Zosa, and Riek}{Woodworth et~al\mbox{.}}{2018}]%
        {woodworth2018preference}
\bibfield{author}{\bibinfo{person}{Bryce Woodworth}, \bibinfo{person}{Francesco Ferrari}, \bibinfo{person}{Teofilo~E Zosa}, {and} \bibinfo{person}{Laurel~D Riek}.} \bibinfo{year}{2018}\natexlab{}.
\newblock \showarticletitle{Preference learning in assistive robotics: {O}bservational repeated inverse reinforcement learning}. In \bibinfo{booktitle}{\emph{Machine learning for healthcare conference}}. PMLR, \bibinfo{pages}{420--439}.
\newblock


\bibitem[\protect\citeauthoryear{Xihan, Mendez, and Hadfield}{Xihan et~al\mbox{.}}{2022}]%
        {xihan2022skill}
\bibfield{author}{\bibinfo{person}{Bian Xihan}, \bibinfo{person}{Oscar Mendez}, {and} \bibinfo{person}{Simon Hadfield}.} \bibinfo{year}{2022}\natexlab{}.
\newblock \showarticletitle{SKILL-IL: {D}isentangling skill and knowledge in multitask imitation learning}. In \bibinfo{booktitle}{\emph{2022 IEEE/RSJ International Conference on Intelligent Robots and Systems (IROS)}}. IEEE, \bibinfo{pages}{7060--7065}.
\newblock


\bibitem[\protect\citeauthoryear{Zar}{Zar}{2005}]%
        {zar2005spearman}
\bibfield{author}{\bibinfo{person}{Jerrold~H Zar}.} \bibinfo{year}{2005}\natexlab{}.
\newblock \bibinfo{booktitle}{\emph{Spearman rank correlation}}. Vol.~\bibinfo{volume}{7}.
\newblock \bibinfo{publisher}{Wiley Online Library}.
\newblock
\showISBNx{9780470011812}


\bibitem[\protect\citeauthoryear{Zhan, Tao, and Cao}{Zhan et~al\mbox{.}}{2021}]%
        {zhan2021human}
\bibfield{author}{\bibinfo{person}{Huixin Zhan}, \bibinfo{person}{Feng Tao}, {and} \bibinfo{person}{Yongcan Cao}.} \bibinfo{year}{2021}\natexlab{}.
\newblock \showarticletitle{Human-guided robot behavior learning: {A} gan-assisted preference-based reinforcement learning approach}.
\newblock \bibinfo{journal}{\emph{IEEE Robotics and Automation Letters}} \bibinfo{volume}{6}, \bibinfo{number}{2} (\bibinfo{year}{2021}), \bibinfo{pages}{3545--3552}.
\newblock


\bibitem[\protect\citeauthoryear{Zhu, Wong, Mandlekar, Mart\'{i}n-Mart\'{i}n, Joshi, Nasiriany, and Zhu}{Zhu et~al\mbox{.}}{2020}]%
        {robosuite2020}
\bibfield{author}{\bibinfo{person}{Yuke Zhu}, \bibinfo{person}{Josiah Wong}, \bibinfo{person}{Ajay Mandlekar}, \bibinfo{person}{Roberto Mart\'{i}n-Mart\'{i}n}, \bibinfo{person}{Abhishek Joshi}, \bibinfo{person}{Soroush Nasiriany}, {and} \bibinfo{person}{Yifeng Zhu}.} \bibinfo{year}{2020}\natexlab{}.
\newblock \showarticletitle{robosuite: {A} Modular Simulation Framework and Benchmark for Robot Learning}.
\newblock \bibinfo{journal}{\emph{arXiv preprint arXiv:2009.12293}} (\bibinfo{year}{2020}).
\newblock


\bibitem[\protect\citeauthoryear{Zolotas and Demiris}{Zolotas and Demiris}{2022}]%
        {zolotas2022disentangled}
\bibfield{author}{\bibinfo{person}{Mark Zolotas} {and} \bibinfo{person}{Yiannis Demiris}.} \bibinfo{year}{2022}\natexlab{}.
\newblock \showarticletitle{Disentangled sequence clustering for human intention inference}. In \bibinfo{booktitle}{\emph{2022 IEEE/RSJ International Conference on Intelligent Robots and Systems (IROS)}}. IEEE, \bibinfo{pages}{9814--9820}.
\newblock


\end{thebibliography}

\reb{

\appendix

\section{Appendix}

In this section, we include additional experiments that compare our proposed approach, PECAN, with previous methods for learning latent representations of robot trajectories and evaluate its performance for varying design parameters such as the number of training demonstrations and the dimensionality of the canonical space. 

\begin{figure}[b]
    \begin{center}
    \includegraphics[width=\columnwidth]{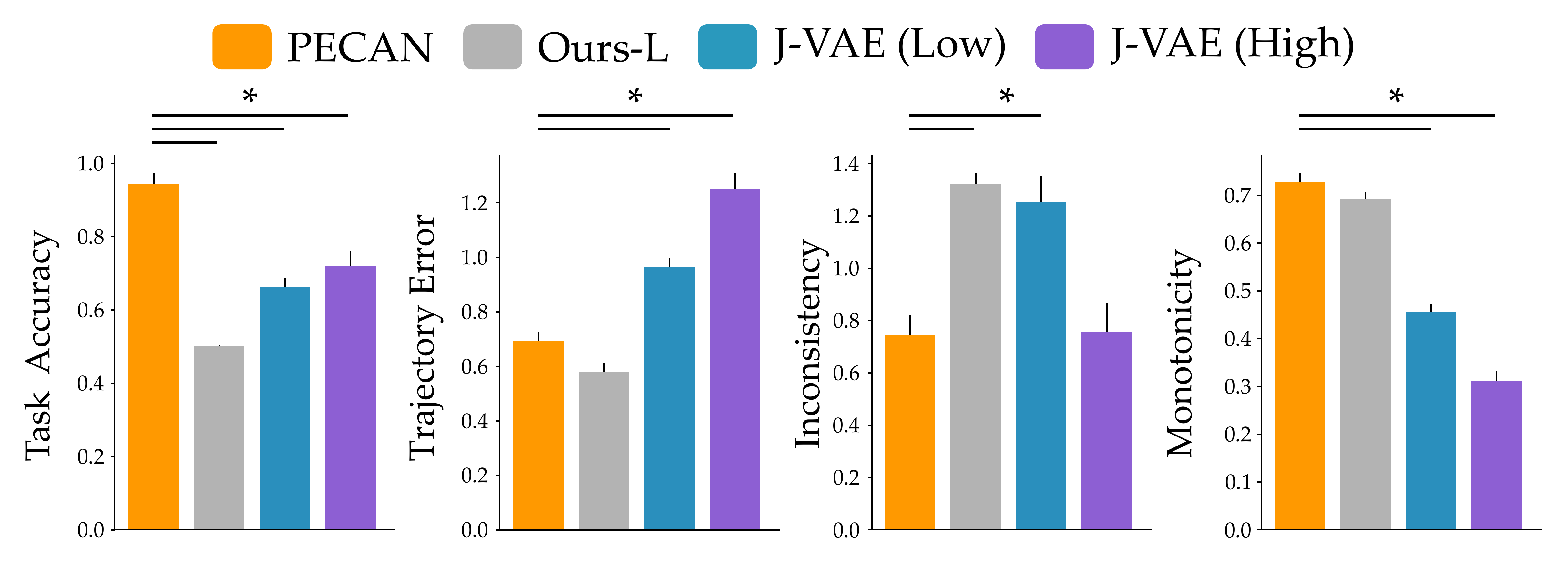}
    \caption{Simulation results comparing PECAN to Joint Variational Autoencoders (VAEs) in the driving environment. We consider two variations of Joint VAEs: (Low) focusing on trajectory reconstruction, and (High) prioritizing separation of tasks and styles. We found that both Joint VAE models had a lower task accuracy and monotonicity than PECAN. While Joint VAE (High) learned consistent canonical spaces similar to PECAN, it had a significantly higher trajectory error. Unlike PECAN, Joint VAEs failed to simultaneously optimize trajectory reconstruction and structuring the latent spaces.
    An asterisk (*) denotes statistical significance and the error bars indicate standard error.}
    \label{fig:rebuttal_baseline}
    \end{center}
\end{figure}

\subsection{Comparing with variational autoencoders (VAEs)} \label{app:baseline}

Previous research has largely focused on learning latent representation for performing downstream robotics tasks. To our knowledge, our approach is the first to explore how these representations can be made more intuitive for humans to directly interact with and modify the robot's behavior. Here we compare PECAN to a state-of-the-art approach for learning latent task and style representations, demonstrating that the representations learned by PECAN are better at exhibiting the user-friendly characteristics of \textit{Consistency} and \textit{Monotonicity} described in Section~\ref{sec:method-characteristics}.

We considered an unsupervised approach~\cite{osa2020goal} that employs \textbf{Joint VAE}s~\cite{dupont2018learning} to learn separate continuous and discrete latent spaces. This approach is comparable to an ablation of our approach that does not incorporate any labels (\textbf{Ours-L}). However, instead of using autoencoders as in our proposed architecture, it leverages VAEs which are widely used for learning latent representations of robot trajectories~\cite{co2018self, allshire2021laser, sun2021task, wang2022skill}. We compare \textbf{PECAN} against two variations of \textbf{Joint VAE}s: the first, \textbf{Joint VAE (Low)}, places less emphasis on structuring the latent spaces and focuses on reconstructing the trajectories, while the second, \textbf{Joint VAE (High)}, prioritizes disentanglement of the latent task and style spaces over trajectory reconstruction. We test these approaches in the driving environment following the same procedure as in our simulation experiments in Section~\ref{sec:simulations}. For a fair comparison, we trained all methods with the same amount of training data and using similar network architectures (number of hidden layers, parameters, activation functions, etc.). The main differences between the methods were the loss functions used to train the networks.

Our results are summarized in \fig{rebuttal_baseline}. We observed that \textbf{PECAN} significantly outperformed both the baselines in accurately encoding the latent tasks and reconstructing the robot trajectories. One-way ANOVA tests revealed that the methods had significant effects on \textit{Task Accuracy} ($F(3, 76)=44.0$, $p<0.01$) and \textit{Trajectory Error} ($F(3, 76)=56.3$, $p<0.01$). Since \textbf{Joint VAE (Low)} prioritized trajectory reconstruction, similar to \textbf{Ours-L}, its canonical spaces were less consistent and monotonic than those learned by \textbf{PECAN}.
Although \textbf{Joint VAE (High)} was able to disentangle the style and task information, and learn consistent canonical spaces, it had a significantly lower monotonicity than PECAN. One-way ANOVA tests also revealed significant effects of the choice of method on \textit{Inconsistency} ($F(3, 76)= 13.1$, $p<0.01$) and \textit{Monotonicity} ($F(3, 76)= 122.0$, $p<0.01$).

\p{Takeaway} Unlike previous approaches that impose priors on the latent spaces to disentangle the task and style information, PECAN leverages labels for trajectories with similar styles to separate the latent tasks and styles. These labels also enable PECAN to learn consistent and monotonic canonical spaces, making them intuitive for users to directly personalize the robot's behavior.

There are several design parameters that enable PECAN to learn a user-friendly representation of styles and accurately produce trajectories for different latent values. In the following sections we will analyze how these parameters impact PECAN's performance.

\subsection{Varying number of latent dimensions}\label{app:dimension}

We start by evaluating PECAN's sensitivity to the dimensionality of the canonical space. In our experiments thus far, we assumed that the dimensionality of the latent styles is known or can be estimated based on the number of extreme labels provided by the user. However, if there is a mismatch between the number of latent dimensions and the dimensions that the user wants to personalize, the learned canonical space may not exhibit the required user-friendly properties. 

We conduct a new experiment in the driving environment to assess PECAN's performance when using under-defined and over-defined canonical spaces. The styles in this environment are two-dimensional, resulting in four style extremes. We thus need a 2D canonical space to encode the styles intuitively. We compared training PECAN using a two-dimensional canonical space (\textit{Exact}) to using canonical spaces with one fewer dimension (\textit{Fewer}) and one extra dimension (\textit{Extra}). 

We followed a similar training and testing procedure as in Section~\ref{sec:simulations}. In addition to the dependent variables from our simulation experiments, we considered a new metric called \textit{Disentanglement}, which evaluates the correlation between individual dimensions of the canonical space and the actual styles. For instance, given a canonical space $Z_{\theta} = [Z_{1}, Z_{2}]$ and robot styles $\theta = [\theta_{1}, \theta_{2}]$, we first list all alignments between the dimensions, e.g., $\left\{[(Z_{1}, \theta_{1}), (Z_{2}, \theta_{2})], [(Z_{1}, \theta_{2}), (Z_{2}, \theta_{1})]\right\}$. Then for each alignment, we compute the mean absolute value of Pearson correlation coefficient $\rho$~\cite{cohen2009pearson} between the paired dimensions:
$$\rho_{avg} = \frac{|\rho_{Z_{1}, \theta_{1}}| + |\rho_{Z_{2}, \theta_{2}}|}{2}$$
Finally, we compute the \textit{Disentanglement} between $Z$ and $\theta$ as the maximum value of $\rho_{avg}$ across all alignments. Unlike \textit{Monotonicity}, which measures whether the differences in latent values are rank correlated to the differences in latent styles, \textit{Disentanglement} measures whether each latent dimension independently captures a distinct style variable.

\fig{rebuttal_dimensionality} summarizes our results averaged over 20 training and testing runs. One way ANOVA tests indicated a significant effect of dimensionality mismatch on all dependent variables: \textit{Task Accuracy} ($F(2, 57)= 93.8$, $p < 0.01$), \textit{Trajectory Error} ($F(2, 57)= 8.61$, $p < 0.01$), \textit{Inconsistency} ($F(2, 57)= 79.5$, $p < 0.01$), \textit{Monotonicity} ($F(2, 57)= 91.8$, $p < 0.01$), and \textit{Disentanglement} ($F(2, 57)= 64.3$, $p < 0.01$). Specifically, we observed that when we added an extra dimension to the canonical space, it inadvertently captured the task information along with the styles. As a result, all trajectories were mapped to the same latent task, significantly reducing the \textit{Task Accuracy} and the \textit{Consistency} of the canonical space. However, the extra dimension allowed our cross-entropy loss in Equation~\ref{eq:cross} to arrange the extreme styles such that the monotonicity of the canonical space was retained. The correlation between individual dimensions (\textit{Disentanglement}) was also preserved because the extra dimension mainly captured the task information, allowing the other two latent dimensions to align with the actual style.
In contrast, when the canonical space had fewer dimensions than the actual style, the accuracy of task encodings was unaffected. Reducing the dimensionality of the canonical space only impacted PECAN's ability to arrange the styles monotonically and disentangle the style dimensions. Since the driving styles were two-dimensional, it was not possible to disentangle them using a 1D canonical space.

\begin{figure}[t]
    \begin{center}
    \includegraphics[width=\columnwidth]{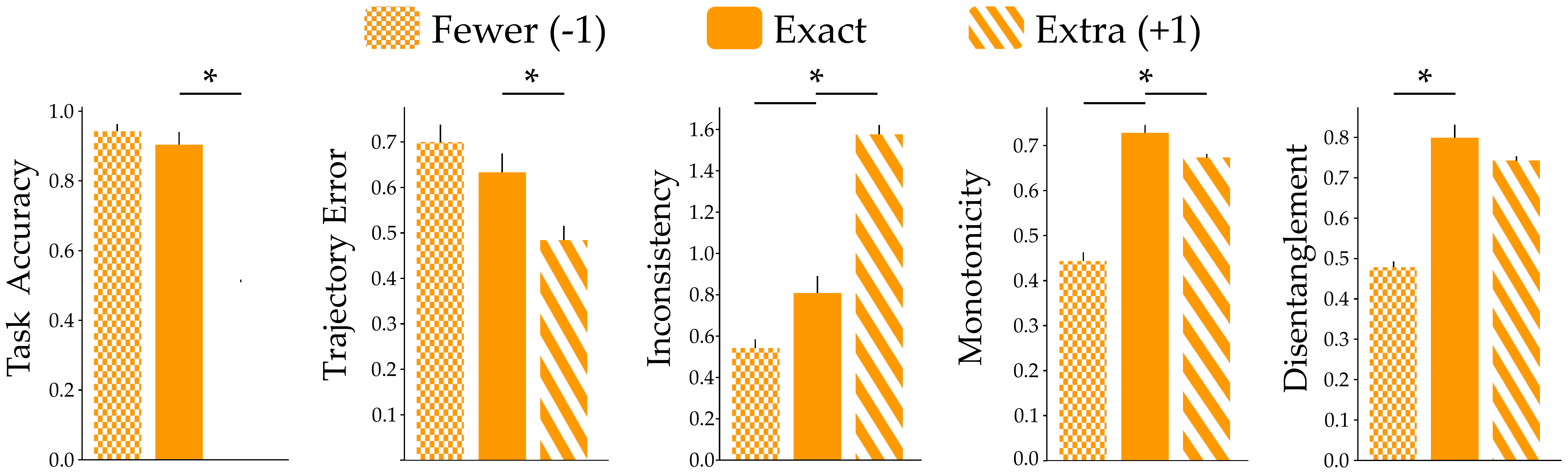}
    \caption{Simulation results for training PECAN with canonical spaces having exact, fewer, and extra dimensions than the true styles. We observed that when the canonical space had an extra dimension, it inadvertently captured the task information, encoding similar trajectories to different latent styles. This reduced the accuracy of PECAN's task encodings as well as the consistency of its canonical space. On the other hand, when the canonical space had one less dimension than the actual styles, it lost its monotonicity because multiple style dimensions were compressed to a single latent dimension
    . An asterisk (*) denotes statistical significance and the error bars indicate standard error.}
    \label{fig:rebuttal_dimensionality}
    \end{center}
\end{figure}

\p{Takeaway} Overall, our results highlight the importance of matching the dimensionality of the canonical space to number of style dimensions that the user wants to personalize. While adding extra latent dimensions decreases PECAN's task accuracy and consistency, reducing the dimensions only affects the monotonicity of its canonical space.

\subsection{Increasing demonstrations of intermediate styles}\label{app:intermediate}

PECAN enables users to personalize the robot's trajectory by clicking on any point in the learned canonical space. To accurately reconstruct trajectories for different latent styles from this space, the training dataset must include sufficient demonstrations representing the intermediate styles. In our experiments and user studies, we trained PECAN on a dataset containing labeled demonstrations of all style extremes and only a few unlabeled demonstrations of intermediate styles. We now evaluate how PECAN's performance varies as we increase the number of intermediate styles in the training dataset while keeping the extreme styles constant. Since the intermediate styles are unlabeled, we do not expect them to impact the user-friendly characteristics of the canonical space.

We compared instances of PECAN trained on datasets containing $16$, $32$, $48$, and $64$ demonstrations in the driving environment following the same training and testing procedure as in Section~\ref{sec:simulations}. Eight demonstrations in each dataset represented the extreme styles, while the remaining demonstrations corresponding to intermediate styles were sampled randomly. \fig{rebuttal_demonstrations} showcases our results averaged over $20$ training and testing runs. As expected, increasing the number of intermediate demonstrations did not interfere with learning a user-friendly canonical space. One-way ANOVA tests indicated that the number of training demonstrations had no significant effect on the \textit{Inconsistency} ($F(3, 76)=1.7$, $p=0.17$) and \textit{Monotonicity} ($F(3, 76)=1.7$, $p=0.16$) of the learned canonical space. On the other hand, providing more demonstrations greatly enhanced the accuracy of PECAN's reconstructions. A one-way ANOVA test revealed a statistically significant effect of dataset size on \textit{Trajectory Error} ($F(3, 76)=26.3$, $p<0.01$). At the same time, we observed a slight decrease in the accuracy of PECAN's task encodings, but this drop was not statistically significant ($F(3, 76)=2.0$, $p=0.12$).

\begin{figure}[t]
    \begin{center}
    \includegraphics[width=\columnwidth]{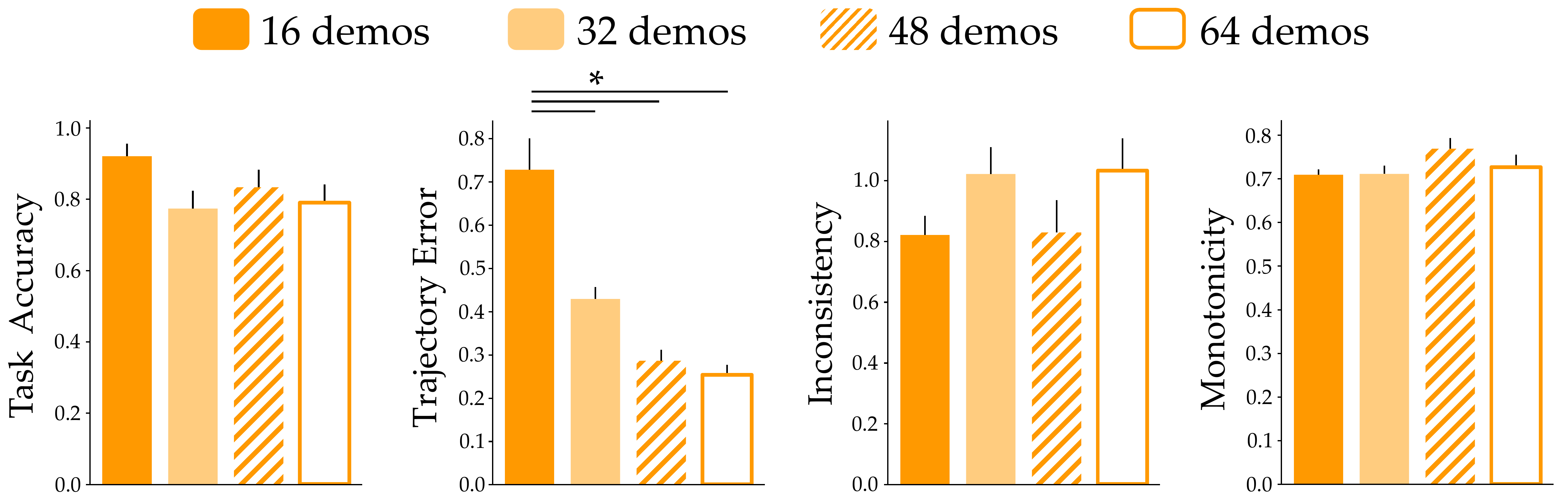}
    \caption{Simulation results for PECAN in the driving environment with increasing demonstrations in the training dataset $|\mathcal{D}| = [16, 32, 48, 64]$. Each dataset contained $8$ demonstrations representing the extreme styles while the remaining demonstrations corresponded to randomly sampled intermediate styles. We observed a significant improvement in the accuracy of PECAN's reconstructions as we increased the number of intermediate styles in the training dataset. Meanwhile, there was no significant change in the accuracy of its task encodings or the consistency and monotonicity of its canonical space. An asterisk (*) denotes statistical significance and the error bars indicate standard error.}
    \label{fig:rebuttal_demonstrations}
    \end{center}
\end{figure}

\p{Takeaway} Overall, these results demonstrate that including more demonstrations of intermediate styles in the training data can improve PECAN's reconstructions, while preserving its user-friendly properties of consistency and monotonicity.

\subsection{Generalizing to higher dimensions}\label{app:generalize}

A key characteristic that makes PECAN's interface user-friendly is the monotonicity of its canonical space. Thus far, we have shown that PECAN can learn monotonic representations for various robot styles ranging from one to three dimensions. Here we test whether PECAN can maintain its user-friendly characteristics when learning high-dimensional canonical spaces.

\begin{figure}[t]
    \begin{center}
    \includegraphics[width=\columnwidth]{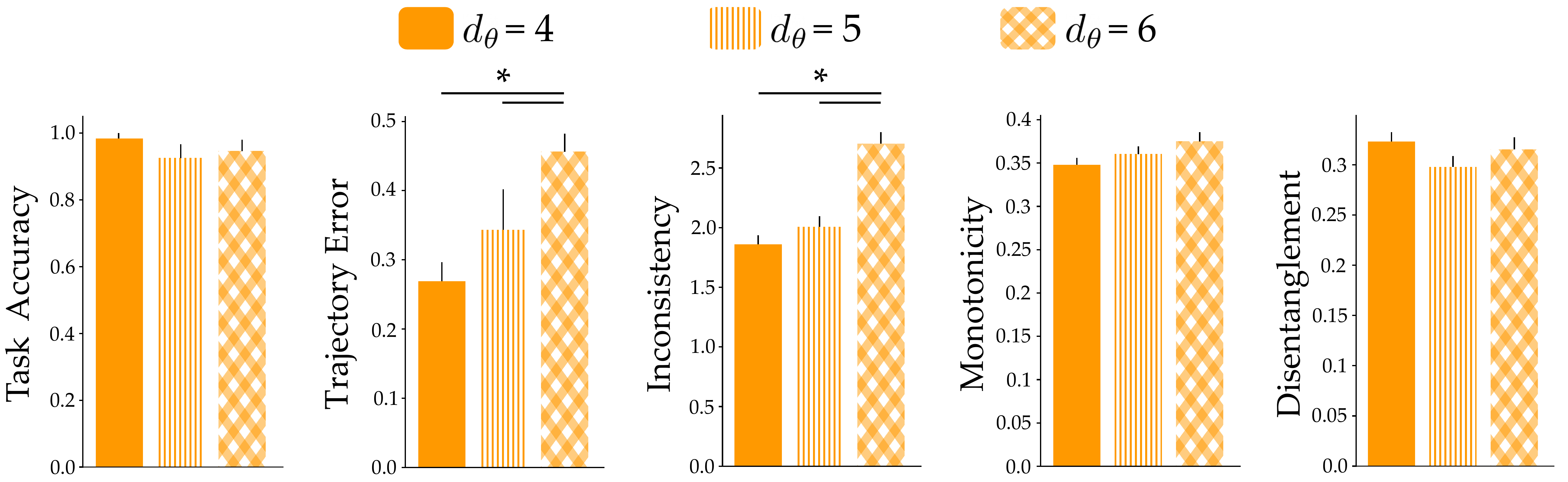}
    \caption{Simulation results for learning high-dimensional canonical spaces. We consider settings in which the robot's trajectory is a polynomial with $d_{\theta}$ parameters that define its style. We compare PECAN's performance for learning styles with $d_{\theta} = [4, 5, 6]$ dimensions. Our results indicate that the canonical spaces learned by PECAN become more inconsistent as we increase the dimensionality of styles. However, this can occur because the size of the canonical space also increases with increasing dimensions. Importantly, we observe that PECAN accurately encodes tasks and maintains monotonicity for high-dimensional canonical spaces. An asterisk (*) denotes statistical significance and the error bars indicate standard error.}
    \label{fig:rebuttal_highdims}
    \end{center}
\end{figure}

We consider a setting where the robot's behavior is defined by an $n$ degree polynomial:
$$y  = \left(a_{0} + a_{1}x + a_{2}x^{2} + \ldots + a_{n}x^{n} \right) \cdot b $$
The setting has two tasks that are determined by $b=\{+1, -1\}$. The robot's trajectory is a sequence of states $(x, y)$, and its style is characterized by the $d_{\theta} = n+1$ polynomial coefficients $[a_{0}, \ldots, a_{n}]$. In this experiment, we learn canonical spaces for styles of three different dimensionalities,  $d_{\theta} \in \{4, 5, 6\}$. For each dimensionality, we first generate a set of $|\Xi| = 3^{d_{\theta}}$ trajectories with $x$ and $a$ sampled uniformly from a fixed range. From this set, we sample a training dataset $\mathcal{D}$ of $|\Xi|/2$ demonstrations, containing $2^{d_{\theta}}$ labeled demonstrations of the style extremes. We use this dataset to train canonical spaces with the same number of dimensions as the actual styles and test the monotonicity of the latent styles on the entire set of trajectories. We compare results for the same dependent variables as in our simulation experiments in Section~\ref{sec:simulations}.

\fig{rebuttal_highdims} summarizes our results averaged over $20$ training and testing runs. 
We found that the canonical spaces became more inconsistent as we increased the style dimensions. A one-way ANOVA test revealed a significant effect of dimensionality on \textit{Inconsistency} ($F(2, 57)=26.5$, $p<0.01$). Pairwise comparisons using Tukey's HSD post-hoc test only indicated a significant drop in consistency at $d_{\theta} = 6$ but not for $d_{\theta} = 5$. This result can be explained by the increasing distance between points (e.g., corners) in a latent space as we add more dimensions. 
While we also observed a decrease in the accuracy of trajectory reconstructions, the \textit{Task Accuracy} was preserved. As demonstrated in Appendix~\ref{app:intermediate}, we can improve PECAN's accuracy by including more intermediate styles in the training demonstrations.
Importantly, we observed that the \textit{Monotonicity} of the canonical space remained consistent despite increasing the dimensionality of the styles. A one-way ANOVA test indicated no significant effect of dimensionality on PECAN's \textit{Monotonicity} ($F(2, 57)=2.0$, $p>0.05$). 

\p{Takeaway} Overall, our results demonstrate that PECAN successfully retains the user-friendly property of \textit{Monotonicity} when learning high-dimensional latent representations of robot styles. While we observe a drop in \textit{Consistency} at higher dimensions, further experiments with real users are necessary to determine whether this result impacts the ease and intuitiveness of using the interface to personalize the robot's style or if it is merely an outcome of computing and comparing distances in high-dimensional spaces.

}

\end{document}